\documentclass[11pt]{article}

\usepackage[final]{acl}
\usepackage{times}
\usepackage{latexsym}
\usepackage[T1]{fontenc}
\usepackage[utf8]{inputenc}
\usepackage{microtype}
\usepackage{inconsolata}
\usepackage{amsmath}
\usepackage{amssymb}
\usepackage{graphicx}
\usepackage{booktabs}
\usepackage{multirow}
\usepackage{xcolor}
\usepackage{hyperref}
\usepackage{subcaption}
\usepackage{enumitem}
\usepackage{tikz}
\usepackage{array}
\usepackage{tabularx}
\usetikzlibrary{shapes,arrows,positioning,fit,backgrounds,calc}

\title{A Survey on Rubric-Guided Reinforcement Learning \\ for Language Models}

\author{Zifei Shan\thanks{\ \ Equal contribution.} \\
  WeChat, Tencent \\
  \texttt{zifeishan@tencent.com} \\\And
  Fangning Shao\footnotemark[1] \\
  Independent Researcher \\
  \texttt{fangning.shao@gmail.com} \\}

\begin{document}
\maketitle

\begin{abstract}
Reinforcement learning from human feedback (RLHF) has become the dominant paradigm for aligning large language models (LLMs) with human preferences. However, traditional RLHF relies on scalar reward signals that lack interpretability and fail to capture the multifaceted nature of response quality. \textit{Rubric-guided reinforcement learning} addresses these limitations by introducing structured, interpretable evaluation criteria, or rubrics, as the backbone of reward design, feedback generation, and policy optimization. In this survey, we introduce a Bayesian framework that defines constitutions as prior distributions $P(\mathcal{R})$ over evaluation criteria and rubrics as conditional instantiations $\mathcal{R}_x \sim P(\mathcal{R}|x)$. Under this unified view, we present a taxonomy of rubric-guided RL along the prior-posterior axis, covering constitutional AI, instance-specific rubrics, process-level supervision, self-evolving rubrics, and their agentic and multimodal extensions. Furthermore, as rubrics are natural-language artifacts, we present a linguistic analysis of how granularity trade-offs, semantic drift, and linguistic reward hacking impact alignment reliability, identifying key open problems for future research.
\end{abstract}

%%%%%%%%%%%%%%
\section{Introduction}
\label{sec:intro}

The alignment of large language models (LLMs) with human values and preferences has emerged as one of the most critical challenges in modern AI research. Reinforcement learning from human feedback (RLHF, \citealp{ouyang2022training}) has established itself as the predominant paradigm for this alignment, demonstrating that models trained with human-preference-driven reward signals can produce more helpful, harmless, and honest outputs. However, as LLMs are deployed in increasingly diverse and complex domains, the limitations of traditional RLHF, which relies on scalar, opaque reward signals, have become increasingly apparent~\cite{gao2023scaling, finegrained2023, armorm2024}.

A \textit{rubric} is a structured set of evaluation criteria that specifies what constitutes different levels of performance across multiple dimensions. Rubrics have found natural application in LLM evaluation and alignment, where they provide interpretable, multi-dimensional, and controllable feedback signals~\cite{hashemi2024llmrubric, rar2025, rlcf2025}. The integration of rubrics into reinforcement learning for language models, which we term \textit{Rubric-guided Reinforcement Learning}---represents a paradigm shift from implicit, scalar reward optimization to explicit, structured, and interpretable alignment~\cite{rar2025, rubricarm2026, openrubrics2025}.

The motivation for rubric-guided RL is threefold. First, \textbf{interpretability}: scalar reward signals in traditional RLHF are opaque, making it difficult to understand why a particular response receives a high or low score; rubric and checklist methods decompose feedback into human-readable criteria~\cite{hashemi2024llmrubric, rlcf2025}. Second, \textbf{controllability}: rubrics allow fine-grained control over which aspects of response quality are rewarded, enabling alignment that respects specific values, safety constraints, or domain requirements~\cite{armorm2024, qalign2025}. Third, \textbf{self-evolution}: rubrics can be generated, refined, and co-evolved with the policy model, enabling continual self-improvement without external supervision, a capability that scalar reward models lack~\cite{evolm2026, rubricarm2026}.

The landscape of structured-feedback alignment has expanded rapidly since 2022. Constitutional AI~\cite{bai2022constitutional} pioneered the use of explicit natural-language principles to guide AI feedback for alignment. Subsequent work has connected rubric-guided alignment to process supervision and verifier-style feedback~\cite{lightman2023lets, rar2025}, RLVR for objectively checkable reasoning tasks~\cite{shao2024deepseekmath, deepseekr12025, qwenmath2024}, multi-objective reward modeling~\cite{armorm2024}, and direct preference optimization~\cite{rafailov2023dpo}.
Furthermore, Rule Based Rewards~\cite{mu2024rule} demonstrated that composable natural-language rules can serve as direct RLHF rewards for safety and controllability.
Recent rubric-specific preprints and workshop papers further explore rubric generation, rubric-based reward modeling, and self-evolving criteria; we discuss these emerging directions in the taxonomy below.

Existing surveys primarily organize the literature around LLM judges, reward design and PRMs, or self-evolution loops~\cite{gu2024survey, rewarddesign2025, prmsurvey2025, tao2024selfevolution}. We instead take explicit structured natural-language criteria as the organizing object across feedback construction, reward modeling, and policy updates. Accordingly, we use a two-tier inclusion rule: methods are \textit{core} when such criteria directly condition the training or evaluation signal; generic DPO, RLVR, PRMs, verifiers, and preference-learning methods are \textit{supporting mechanisms} unless explicitly rubric-conditioned. This boundary also motivates our linguistic analysis rather than treating rubrics only as optimization or prompting components.

We introduce a \textbf{Bayesian framework} for rubric-guided RL: constitutions are \textit{prior distributions} $P(\mathcal{R})$ over evaluation criteria (global, input-independent), while rubrics are \textit{conditional instantiations} $\mathcal{R}_x \sim P(\mathcal{R}|x)$ (input-specific). This analytical lens places fixed, conditional, and update-oriented criteria in a common vocabulary and interprets rubric exploitation as overfitting to a point estimate; it does not imply that every self-evolving system performs exact Bayesian updating.

We also foreground a \textbf{linguistic analysis} of rubrics as natural language texts (\S\ref{sec:linguistic}), examining how semantic drift, granularity effects, ambiguity, and linguistic reward hacking affect rubric-guided RL. This NLP-centered perspective is essential because rubrics are fundamentally natural language artifacts, and their textual properties have been underexplored relative to their mathematical formalization.

Our contributions are as follows:
\begin{enumerate}[leftmargin=*]

    \item We introduce a \textbf{Bayesian framework} for rubric-guided RL, defining constitutions as prior distributions $P(\mathcal{R})$ over evaluation criteria and rubrics as conditional instantiations $P(\mathcal{R}|x)$. This framework unifies fixed constitutions, instance-specific rubrics, and self-evolving criteria, while clarifying its scope and limitations (\S\ref{sec:framework}).

    \item We provide a \textbf{linguistic analysis} of rubrics as natural-language artifacts, examining semantic drift, granularity effects, ambiguity, and linguistic reward hacking---phenomena that directly affect the stability and reliability of rubric-guided RL but remain underdeveloped in prior work (\S\ref{sec:linguistic}).

    \item We synthesize a \textbf{prior-posterior taxonomy} of rubric-guided RL, covering fixed constitutional priors, deterministic instance-specific rubrics,
    process-level supervision, rubric-based reward modeling, self-evolving rubrics, and extensions to agentic and multimodal settings.

\end{enumerate}

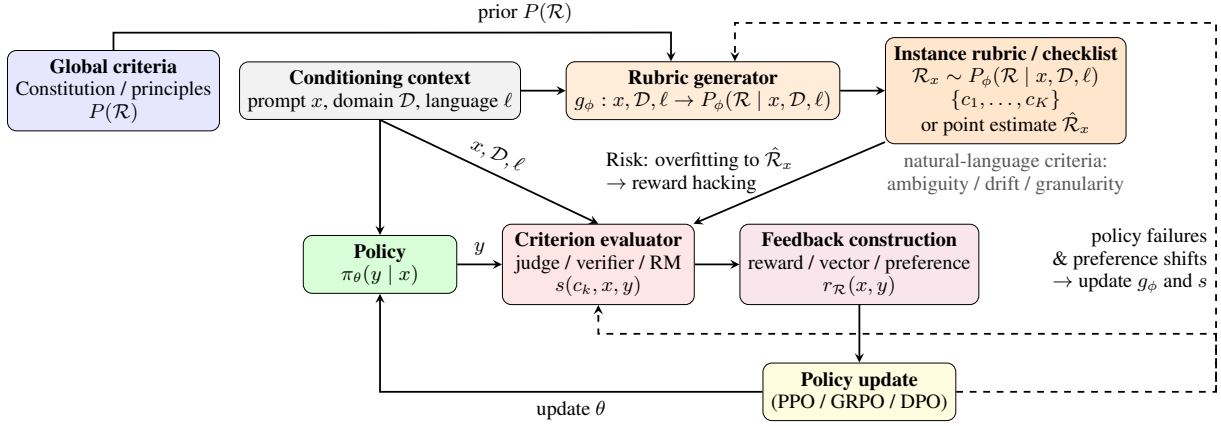
\begin{figure*}[ht]
\centering
\resizebox{\linewidth}{!}{%
\begin{tikzpicture}[
    node distance=1.0cm and 0.7cm,
    block/.style={rectangle, draw, rounded corners, minimum height=0.9cm, minimum width=2.4cm, align=center, font=\small},
    wideblock/.style={rectangle, draw, rounded corners, minimum height=0.9cm, minimum width=3.0cm, align=center, font=\small},
    arrow/.style={->, >=stealth, thick},
    feedback_arrow/.style={->, >=stealth, thick, dashed},
    context_arrow/.style={->, >=stealth, thick},
    note/.style={font=\footnotesize, text=gray!70!black, align=center},
]

% Top row: Bayesian construction of criteria
\node[wideblock, fill=blue!10] (prior) {\textbf{Global criteria}\\Constitution / principles\\$P(\mathcal{R})$};

\node[wideblock, fill=gray!10, right=0.3cm of prior] (context) {\textbf{Conditioning context}\\prompt $x$, domain $\mathcal{D}$, language $\ell$};

\node[wideblock, fill=orange!15, right=of context] (generator) {\textbf{Rubric generator}\\$g_\phi:x,\mathcal D,\ell \rightarrow P_\phi(\mathcal R\mid x,\mathcal D,\ell)$};

\node[wideblock, fill=orange!20, right=of generator] (posterior) {\textbf{Instance rubric / checklist}\\$\mathcal R_x \sim P_\phi(\mathcal R\mid x,\mathcal D,\ell)$\\$\{c_1,\ldots,c_K\}$\\or point estimate $\hat{\mathcal{R}}_x$};

% Bottom row: reward construction and optimization
\node[block, fill=green!15, below=1.8cm of context] (policy) {\textbf{Policy}\\$\pi_\theta(y\mid x)$};

\node[wideblock, fill=red!10, right=of policy] (judge) {\textbf{Criterion evaluator}\\judge / verifier / RM\\$s(c_k,x,y)$};

\node[wideblock, fill=purple!10, right=of judge] (agg) {\textbf{Feedback construction}\\reward / vector / preference\\$r_{\mathcal{R}}(x, y)$ };

\node[wideblock, fill=yellow!15, below=0.9cm of agg] (opt) {\textbf{Policy update}\\(PPO / GRPO / DPO)};

% ==================== Main arrows ====================
\draw[arrow] (prior.north) |- ([yshift=0.45cm]context.north)
    -- ([xshift=-0.5cm, yshift=0.45cm]generator.north) node[midway, above, font=\footnotesize] {prior $P(\mathcal{R})$}
    -- ([xshift=-0.5cm]generator.north);

\draw[arrow] (context) -- (generator);
\draw[arrow] (generator) -- (posterior);
\draw[arrow] (context) -- (policy);

\draw[context_arrow] (context.south) -- (judge.north) node[midway, above, sloped, font=\footnotesize] {$x,\mathcal{D},\ell$};

\draw[arrow] (policy) -- (judge) node[midway, above, font=\footnotesize] {$y$};

\draw[arrow] (posterior) -- (judge) node[midway, left=-0.2cm, yshift=0.2cm, font=\footnotesize, align=left] {Risk: overfitting to $\hat{\mathcal{R}}_x$\\$\rightarrow$ reward hacking};

\draw[arrow] (judge) -- (agg);
\draw[arrow] (agg) -- (opt);
\draw[arrow] (opt.west) -| (policy.south) node[pos=0.25, below, font=\footnotesize] {update $\theta$};

% ==================== Feedback Loop ====================
\draw[feedback_arrow] (opt.east)
    -| ([xshift=1.4cm]posterior.east) coordinate (vpoint)
    |- ([xshift=0.5cm, yshift=0.6cm]generator.north)
    -- ([xshift=0.5cm]generator.north);

\draw[feedback_arrow] (opt.east -| vpoint)
    |- ([yshift=-0.45cm]judge.south)
    -- (judge.south);

\path (opt.east -| vpoint) -- (vpoint) node[midway, left, yshift=-0.3cm, font=\footnotesize, align=right] {policy failures \\ \& preference shifts \\ $\rightarrow$ update $g_\phi$ and $s$};

% ==================== Annotations ====================
\node[note, below=0.01cm of posterior] {natural-language criteria:\\ambiguity / drift / granularity};

\end{tikzpicture}%
}
\caption{
Bayesian view of rubric-guided RL. Global principles define a prior
$P(\mathcal{R})$ over evaluation criteria. Given $x$, $\mathcal{D}$, and $\ell$,
the rubric generator $g_\phi$ parameterizes $P_\phi(\mathcal{R}\mid x,\mathcal{D},\ell)$
and produces sampled rubrics $\mathcal{R}_x$ or deterministic estimates
$\hat{\mathcal{R}}_x$. Criterion evaluators construct scalar, vector, or preference
feedback for policy optimization. Dashed arrows indicate self-evolution of the
generator or evaluator; reward hacking corresponds to overfitting to brittle
rubric point estimates.
}
\label{fig:architecture}
\end{figure*}

%%%%%%%%%%%%%%
\section{Rubric: Definition and Formalization}
\label{sec:rubric-def}

In this section, we define what a rubric is, formalize its role as a reward signal, and introduce a unifying Bayesian framework that clarifies the relationship between rubrics, constitutions, and principles.

\subsection{What is a Rubric?}

A rubric is a structured evaluation instrument that specifies criteria for assessing performance across multiple dimensions and quality levels. In the context of LLM alignment, a rubric $\mathcal{R}$ for a given prompt $x$ can be formalized as a set of criteria:
$\mathcal{R}(x) = \{c_1, c_2, \ldots, c_K\}$,
where each criterion $c_k$ specifies an evaluation dimension (e.g., correctness, completeness, clarity, safety) and may include performance-level descriptors. Unlike scalar reward signals, rubrics decompose the evaluation into interpretable, auditable components. Importantly, rubrics are \textit{natural language texts}: their linguistic properties, including ambiguity, granularity, and semantic stability, fundamentally affect rubric-guided RL performance, as analyzed in \S\ref{sec:linguistic}.

\subsection{Rubric as Reward Signal}

The key insight of rubric-guided RL is that rubrics can serve as structured reward signals. Given a rubric $\mathcal{R}(x) = \{c_1, \ldots, c_K\}$ and a response $y$, the rubric-conditioned reward can be computed as:
$r_{\mathcal{R}}(x, y) = \sum_{k=1}^{K} w_k \cdot s(c_k, x, y)$,
where $w_k$ is the weight for criterion $c_k$ and $s(c_k, x, y)$ is the criterion-specific score. This formulation enables interpretable, controllable, and multi-dimensional reward computation.

\subsection{Rubric Generation}

Rubric generation is a critical component of rubric-guided RL. Approaches span from contrastive rubric generation (CRG)~\cite{openrubrics2025}, instance-level checklist generation~\cite{rlcf2025}, and rubric generation for judges~\cite{rrd2026, autorubric2026} to self-proposed rubrics~\cite{rlcer2026, evolm2026} and fixed constitutional principles~\cite{bai2022constitutional}. The Bayesian framework distinguishes methods that optimize distributions over rubrics from those that construct deterministic, input-conditioned criteria.

\subsection{A Bayesian Framework: Constitution as Prior, Rubric as Posterior}
\label{sec:framework}

To provide a unifying perspective for rubric-guided RL, we introduce a \textit{Bayesian framework} that offers a probabilistic notation for organizing how different methods specify, condition, and update evaluation criteria. This framework resolves terminological ambiguity by explicitly clarifying the relationship between constitutions, rubrics, and principles:

\begin{itemize}[leftmargin=*]
    \item A \textbf{constitution} $P(\mathcal{R})$ functions as a
    \textit{prior distribution} over evaluation criteria: a global,
    input-independent specification of what quality dimensions are relevant.
    Constitutional AI~\cite{bai2022constitutional} defines such a prior through
    natural-language principles (e.g., helpfulness, harmlessness, honesty) that
    apply across inputs. We classify constitutions as priors based on their
    functional role: they specify global evaluation criteria before observing a
    specific input.

    \item The conditional distribution $P(\mathcal{R}|x)$ functions as a
    \textit{posterior distribution} over evaluation criteria, conditioned on the
    observed input $x$. An instance-specific \textbf{rubric} $\mathcal{R}_x$ is a
    sampled or deterministic instantiation of this posterior. Instance-specific
    methods such as RaR~\cite{rar2025} approximate $P(\mathcal{R}|x)$ with a
    point estimate or checklist, while domain-specific rubrics correspond to
    coarser conditionals such as $P(\mathcal{R}|\mathcal{D})$. General rubrics
    are closest to the prior $P(\mathcal{R})$ with minimal conditioning. The
    Bayesian framing eliminates the need for a rigid hierarchy: constitutions and
    rubrics can be viewed as the same evaluation object at different levels of
    conditioning.

    \item A \textbf{principle} is an atomic element of the evaluation space: a
    single dimension along which quality can be assessed.
    % Principles form the
    % support of both the prior $P(\mathcal{R})$ and the posterior
    % $P(\mathcal{R}|x)$.
    The sample space of the prior $P(\mathcal{R})$ consists of valid combinations of these principles.
    In Constitutional AI~\cite{bai2022constitutional}, each constitutional principle
    is one such dimension.
    % ; in RaR~\cite{rar2025}, each checklist item can be viewed as a principle.
\end{itemize}

This Bayesian framework formalizes constitutions and rubrics as the same mathematical object at different conditioning levels. The constitution is
the unconditional prior $P(\mathcal{R})$; domain-specific rubrics instantiate or
approximate coarser conditionals such as $P(\mathcal{R}|\mathcal{D})$; and
instance-specific rubrics instantiate or approximate the full posterior
$P(\mathcal{R}|x)$.
Treating the input $x$ or domain $\mathcal{D}$ as the observation explains the structural difference in generality between these evaluation instruments.

\paragraph{Formalization.}
We define a rubric-guided RL system as a tuple $(\pi_\theta, P(\mathcal{R}), g_\phi, \mathcal{J}, \mathcal{A})$ where:
\begin{itemize}[leftmargin=*]
    \item $\pi_\theta(y|x)$ is the policy model;
    \item $P(\mathcal{R})$ is the prior over evaluation criteria (the constitution);
    \item $g_\phi$ is the rubric generator that maps conditioning variables such as the prompt $x$, domain $\mathcal{D}$, and language $\ell$ to a conditional distribution $P_\phi(\mathcal{R}\mid x,\mathcal{D},\ell)$, or to a deterministic point estimate $\hat{\mathcal{R}}_x$;
    \item $\mathcal{J}$ is a criterion evaluator, such as an LLM judge, verifier, or reward model, that produces criterion-level scores $s(c_k,x,y)$;
    \item $\mathcal{A}$ constructs feedback from criterion-level scores, such as scalar rewards, multi-dimensional vectors, or preference labels.
\end{itemize}

Appendix~\ref{sec:bayesian-clarification} further clarifies how this notation applies to fixed, deterministic, and self-evolving rubric systems.

Given conditioning variables $(x,\mathcal{D},\ell)$, an explicitly probabilistic rubric generator samples $\mathcal{R}_x \sim P_\phi(\mathcal{R}\mid x,\mathcal{D},\ell)$. Deterministic systems instead return a point estimate $\hat{\mathcal{R}}_x = g_\phi(x,\mathcal{D},\ell)$.

For scalar-reward methods, the rubric-guided reward aggregates criterion-level scores, potentially over multiple rubric samples:
\begin{equation}
\label{eq:expected-reward}
r_{\mathcal{R}}(x, y) = \mathbb{E}_{\mathcal{R}_x \sim P_\phi(\mathcal{R}\mid x,\mathcal{D},\ell)} \Big[\mathcal{A}\big(\{s(c_k, x, y)\}_{k=1}^K\big)\Big]
\end{equation}
These scalar rewards are typically optimized using KL-regularized policy gradients:
\begin{equation}
\label{eq:unified-objective}
\max_\theta \; \mathbb{E}_{x \sim \mathcal{D}, y \sim \pi_\theta}\big[r_{\mathcal{R}}(x, y)\big] - \beta \cdot \text{KL}(\pi_\theta \| \pi_{\text{ref}})
\end{equation}

Alternatively, direct preference optimization (DPO) methods instantiate the same conceptual framework by modifying both endpoints: the aggregation function $\mathcal{A}$ produces pairwise preferences or labels rather than scalar rewards, and the policy is updated via a pairwise preference loss rather than Equation~\ref{eq:unified-objective}.

%%%%%%%%%%%%%%
\section{Taxonomy of Rubric-Guided RL Methods}
\label{sec:taxonomy}

Building on the Bayesian framework introduced in \S\ref{sec:framework}, we categorize existing methods under this lens. Table~\ref{tab:taxonomy-v2} organizes these methods along the prior-posterior axis. Methods differ primarily in how they specify, approximate, or operationalize the transition from global evaluation principles $P(\mathcal{R})$ to input-conditioned criteria $P(\mathcal{R}|x)$. Our coverage prioritizes peer-reviewed publications and stable archival records, while additionally including influential preprints and technical reports that have been widely adopted or cited in the community. Appendix~\ref{sec:survey-methodology} summarizes the search and inclusion protocol.

\begin{table*}[ht]
\centering
\small
\setlength{\tabcolsep}{4pt}
\begin{tabularx}{\linewidth}{@{}l l >{\raggedright\arraybackslash}X >{\raggedright\arraybackslash}X >{\raggedright\arraybackslash}X >{\raggedright\arraybackslash}X@{}}
\toprule
\textbf{Method} & \textbf{Domain} & \textbf{Prior $P(\mathcal{R})$} & \textbf{Posterior $P(\mathcal{R}|x)$} & \textbf{Aggregation} & \textbf{Training / Optim.} \\
\midrule
RaR~\cite{rar2025} & Text & Implicit & Deterministic point & Weighted sum / Implicit & GRPO \\
Rubric-ARM~\cite{rubricarm2026} & Text & Learned & Variational & Learned & PPO \\
EvoLM~\cite{evolm2026} & Text & Implicit & Variational (temporal) & Discriminative & GRPO \\
RLCER~\cite{rlcer2026} & Reasoning & Implicit & Process-conditional & Process-level & Extended from PPO \\
Constitutional AI~\cite{bai2022constitutional} & Text & Fixed principles & Point mass at prior & Principle ensembling & PPO \\
QA-LIGN~\cite{qalign2025} & Text & Fixed principles & QA decomposition & Decomposed & GRPO \\
PRM~\cite{lightman2023lets} & Math/Code & Correctness & Implicit & Step product & RM training \\
RLCF~\cite{rlcf2025} & Text & Implicit & Instance checklist & Weighted sum & DPO \\
OpenRubrics~\cite{openrubrics2025} & Text & CRG & Contrastive & Multi-dimensional & RM training \\
RRD~\cite{rrd2026} & Text & Coarse rubrics & Refined criteria & Corr.-aware weights & RM training \\
OpenRS~\cite{openrs2026} & Text & Principles & Pairwise adaptive & Criterion prefs. & GRPO \\
\midrule
\multicolumn{6}{l}{\textit{Agentic Extensions (see \S\ref{sec:emerging-domains}):}} \\
DR Tulu~\cite{drtulu2025} & Agent & Evolving & Evolving & Evolving & GRPO \\
HiPER~\cite{hiper2026} & Agent & Hierarchical & Plan-Execute & HAE & PPO \\
RubricEM~\cite{rubricem2026} & Agent & Stagewise & Stagewise & Stagewise & Extended from GRPO \\
AdaRubric~\cite{adarubric2026} & Agent & Task desc. & Task-adaptive & Conf.-weighted & DPO \\
\midrule
\multicolumn{6}{l}{\textit{Multimodal Extensions (see \S\ref{sec:emerging-domains}):}} \\
ARR~\cite{arr2026} & VLM & Implicit & Generated & Explicit & GRPO \\
DeltaRubric~\cite{deltarubric2026} & VLM & Implicit & Plan-checklist & Verification & GRPO \\
rDPO~\cite{rdpo2026} & VLM & Implicit & Instance-specific & Extended from DPO \\
Step-Audio-R1.5~\cite{stepaudior152026} & Audio & Implicit & Acoustic criteria & Preference RM & GRPO \\
\bottomrule
\end{tabularx}
\caption{Taxonomy of rubric-guided RL methods under the Bayesian framework. Each method is characterized by its domain, prior specification, posterior parameterization or criterion-construction mechanism, aggregation strategy, and optimization algorithm.}
\label{tab:taxonomy-v2}
\end{table*}

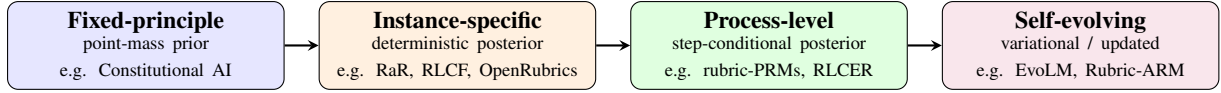
\begin{figure*}[t]
\centering
\begin{tikzpicture}[
    axisblock/.style={rectangle, draw, rounded corners, minimum height=0.75cm, text width=3.4cm, align=center, font=\small},
    axisarrow/.style={->, >=stealth, thick},
    node distance=0.45cm
]
\node[axisblock, fill=blue!10] (fixed) {\textbf{Fixed-principle}\\\scriptsize point-mass prior\\\scriptsize e.g. Constitutional AI};
\node[axisblock, fill=orange!12, right=of fixed] (instance) {\textbf{Instance-specific}\\\scriptsize deterministic posterior\\\scriptsize e.g. RaR, RLCF, OpenRubrics};
\node[axisblock, fill=green!12, right=of instance] (process) {\textbf{Process-level}\\\scriptsize step-conditional posterior\\\scriptsize e.g. rubric-PRMs, RLCER};
\node[axisblock, fill=purple!10, right=of process] (evolving) {\textbf{Self-evolving}\\\scriptsize variational / updated\\\scriptsize e.g. EvoLM, Rubric-ARM};
\draw[axisarrow] (fixed) -- (instance);
\draw[axisarrow] (instance) -- (process);
\draw[axisarrow] (process) -- (evolving);
\end{tikzpicture}
\caption{Navigation along the prior--posterior axis. The axis summarizes how method classes condition or update criteria; it does not assert that every class performs exact Bayesian inference.}
\label{fig:taxonomy-axis}
\end{figure*}

\begin{table*}[t]
\centering
\scriptsize
\setlength{\tabcolsep}{3.5pt}
\renewcommand{\arraystretch}{1.08}
\begin{tabularx}{\linewidth}{@{}l l >{\raggedright\arraybackslash}p{3.4cm} >{\raggedright\arraybackslash}X@{}}
\toprule
\textbf{Method class} & \textbf{Example} & \textbf{Criteria source / update} & \textbf{Workflow into optimization} \\
\midrule
Fixed-principle & Constitutional AI & Global principles or constitution & principles $\rightarrow$ critique/preferences $\rightarrow$ scalar or preference feedback $\rightarrow$ policy optimization \\
Instance-specific & RaR-style methods & Input- or domain-conditioned rubrics/checklists & rubric generation $\rightarrow$ criterion verification $\rightarrow$ aggregated scalar/vector/preference feedback $\rightarrow$ PPO/GRPO/DPO \\
Process-level & RLCER, rubric-conditioned PRMs & Step criteria derived from reasoning traces & trace $\rightarrow$ step decomposition $\rightarrow$ step scoring $\rightarrow$ dense process feedback $\rightarrow$ policy update \\
Self-evolving & EvoLM-style systems & Criteria updated from outputs, failures, or preference shifts & initial criteria $\rightarrow$ policy evaluation $\rightarrow$ rubric/evaluator sharpening $\rightarrow$ subsequent policy optimization \\
\bottomrule
\end{tabularx}
\caption{Operational workflows for the major method classes. Generic optimization backends count as rubric-guided only when their feedback is conditioned on structured criteria.}
\label{tab:workflows}
\end{table*}

\subsection{Point-Mass Priors: Constitutional AI}

Methods in this category use a fixed set of principles as the prior $P(\mathcal{R})$ and do not condition heavily on the input, yielding a posterior that is a point mass at the prior:
$P(\mathcal{R}|x) = \delta_{\mathcal{R}_{\text{constitution}}}(\mathcal{R})$.
Representative works include Constitutional AI~\cite{bai2022constitutional}, IterAlign~\cite{iteralign2024}, ICAI~\cite{icai2024}, and Deliberative Alignment~\cite{deliberative2024}, which apply fixed principle sets across all inputs rather than generating task-specific criteria.

\subsection{Deterministic Posteriors: Instance-Specific Rubrics}

Methods in this category generate instance-specific rubrics deterministically, producing a single point estimate from $P(\mathcal{R}|x)$ without uncertainty quantification, including RaR~\cite{rar2025}, RLCF~\cite{rlcf2025}, rubric-conditioned PRMs~\cite{lightman2023lets}, OpenRubrics~\cite{openrubrics2025}, rubric-generation methods for LLM judges~\cite{rrd2026, autorubric2026, learntojudge2026}, and Critique Models~\cite{critiquellm2024}, which map inputs to specific evaluation structures for fine-grained feedback.

\subsection{Variational Posteriors: Self-Evolving Rubrics}

Self-evolving rubrics occupy the update-oriented end of Figure~\ref{fig:taxonomy-axis}: as the policy $\pi_\theta$ improves, preference data and model outputs shift, and criteria or evaluators are updated sequentially. This paradigm is deeply connected to self-rewarding and meta-rewarding language models~\cite{yuan2024selfrewarding, wu2025metarewarding}, which iteratively improve their own feedback generation, and recent methods including Rubric-ARM~\cite{rubricarm2026}, EvoLM~\cite{evolm2026}, and RLCER~\cite{rlcer2026}. Across these methods, we identify three recurring dynamics: (1) \textit{Posterior sharpening} in explicitly latent-rubric formulations; (2) \textit{Prior simplification}, where brittle gating mechanisms simplify for better generalization; and (3) \textit{Conditional expansion}, where rubrics evolve from outcome-focused to process-focused.

\paragraph{Self-evolution versus posterior updating.}
Crucially, self-evolution is broader than exact Bayesian posterior updating. EvoLM provides a relatively concrete posterior-sharpening interpretation: preference pairs constructed from policy outputs act as observations, and a rubric-conditioned frozen judge's score margin between preferred and dispreferred responses supplies a likelihood-like signal~\cite{evolm2026}. By contrast, systems without an explicit prior, likelihood, and observation model, including RLCER, are update-oriented analogues rather than exact Bayesian inference methods.

\paragraph{Assumptions and limitations by class.}
Fixed-principle methods assume that global principles transfer across inputs; instance-specific methods depend on both rubric-generator and evaluator reliability. Process-level methods additionally assume a meaningful decomposition into steps. Self-evolving methods can adapt to policy shifts, but coupled rubric--policy updates risk feedback loops, criterion drift, and evaluator dependence. Figure \ref{fig:taxonomy-axis} and Table~\ref{tab:workflows} make these operational differences explicit.

%%%%%%%%%%%%%%
\section{Linguistic Insights: Rubrics as Natural Language}
\label{sec:linguistic}

A fundamental but underexplored aspect of rubric-guided RL is that rubrics are \textit{natural language texts}. This section analyzes how their prompt engineering, semantic stability, granularity, ambiguity, and cross-linguistic variation affect
the performance and reliability of rubric-guided RL systems.

\subsection{Rubrics as Prompted Specifications}

When rubrics are used to guide LLM judges or reward models, they become prompted specifications: their wording, formatting, and output schema can affect the resulting scores,
as observed in LLM-as-a-judge frameworks like Prometheus, which train evaluators on customized score rubrics~\cite{kim2023prometheus, kim2024prometheus2}, and FLASK, which decomposes evaluation into fine-grained alignment skills~\cite{ye2024flask}.
Different formats expose different tradeoffs: binary checklists are easy to aggregate but coarse~\cite{rar2025}, free-text critiques are expressive but difficult to scalarize~\cite{critiquellm2024}, and multi-level descriptors offer a middle ground between discriminative power and calibration~\cite{hashemi2024llmrubric}.

\subsection{Granularity Effects}

The granularity of rubric criteria, whether defined over tokens, reasoning steps, whole responses, or decomposed checklist items, affects both evaluation quality and the informativeness of reward signals.
The granularity-density tradeoff mirrors the exploration-exploitation tradeoff in RL: denser rewards enable faster learning but may miss global quality, while sparser rewards capture holistic quality but slow convergence. Existing systems often balance this tradeoff through hybrid designs: PRMs combine dense step-level supervision with outcome-level evaluation~\cite{lightman2023lets}, RLCF combines instance-specific checklists with universal criteria and verifier programs~\cite{rlcf2025}, and OpenRS combines pairwise adaptive meta-rubrics with pointwise verifiable rubrics for guardrails~\cite{openrs2026}.

\paragraph{Lexical and Token-Level Signals.}
At the finest scale, alignment systems may use lexical constraints or token-level shaping terms, such as PPO-style per-token KL penalties, length penalties, keyword inclusion/exclusion checks, profanity filters, or exact-format constraints. Lexical and formatting signals directly enforce observable surface properties, and length- or style-correlated rewards can favor well-formatted or verbose outputs without corresponding improvements in semantic quality~\cite{zeng2024token, gao2023scaling, zheng2023secrets}. Therefore these dense signals should be clearly distinguished from process-level credit assignment strategies.

\paragraph{Sentence/Step-Level Criteria.}
Process-level rubrics (RLCER, \citealp{rlcer2026}; PRMs, \citealp{lightman2023lets}) evaluate intermediate reasoning steps, providing denser reward signals than outcome-level rubrics. PRMs operate at the step level, providing per-step correctness signals and achieving 78.2\% accuracy on MATH~\cite{lightman2023lets}. However, step-level evaluation requires defining what constitutes a ``step'', which is itself a non-trivial decision. Recent self-evolving rubric work suggests that optimal granularity interacts with evaluator capacity: smaller evaluators may require more precise rubrics, whereas larger evaluators can tolerate coarser specifications~\cite{evolm2026}.

\paragraph{Response-Level and Document-Level Criteria.}
Outcome-level rubrics, such as RaR~\cite{rar2025} and Constitutional AI~\cite{bai2022constitutional}, evaluate the complete response. They provide holistic assessment of dimensions such as helpfulness, safety, completeness, and coherence, but the reward is sparse and may obscure which part of the response caused success or failure.

\paragraph{Atomic Checklist Criteria.}
A separate form of granularity comes from decomposing a holistic objective into atomic checklist items rather than changing the textual span of evaluation. RLCF~\cite{rlcf2025}, for example, decomposes tasks into atomic checklist items, illustrating how fine-grained criteria can provide denser feedback for robustness.
Similarly, CheckEval~\cite{lee2025checkeval} demonstrates that Boolean checklist-style decomposition improves evaluator reliability and robustness.
This differs from word-level or sentence-level evaluation: the granularity lies in the criterion decomposition, not necessarily in the textual unit being scored.

\subsection{Semantic Drift in Rubric Evaluation}

\textit{Semantic drift} refers to the phenomenon where the meaning of rubric criteria shifts across RL training iterations, even when the rubric text remains unchanged. This drift occurs because the policy model's interpretation of rubric criteria changes as it adapts to the reward signal. We distinguish semantic drift from the broader concept of concept drift~\cite{gama2014survey} in machine learning: semantic drift is specifically about the \textit{semantic instability} of natural language criteria as they are applied by changing models across training, rather than changes in the data distribution itself.

\paragraph{Causes of Semantic Drift.}
We identify three causes: (1) \textbf{Distribution shift}: as the policy improves, the response distribution changes, altering the context in which rubric criteria are applied. A criterion like ``factual accuracy'' means something different for a model that produces mostly correct responses vs.\ one that produces mostly incorrect responses. EvoLM~\cite{evolm2026} provides direct evidence: rubrics evolve from vague criteria (e.g., ``Ensure mathematical accuracy'') to specific, verifiable facts (e.g., ``The answer is 144, derived from perimeter 48'') as the policy distribution shifts during training. (2) \textbf{Criterion gaming}: the policy learns to satisfy the letter of rubric criteria while violating their spirit, effectively shifting the rubric's semantics. Constitutional AI~\cite{bai2022constitutional} observed this Goodharting effect, where over-optimization leads to boilerplate language and overly harsh responses. (3) \textbf{Evolving evaluator}: in systems with learned judges, reward models, or co-evolving rubric generators, the scoring function itself changes during training. Thus, even when a criterion such as ``mathematical accuracy'' is textually stable, the function $s_t(c_k,x,y)$ that applies this criterion may shift as the evaluator is updated on new policy outputs. This is distinct from judge collapse: evaluator evolution is a general design property of adaptive systems, whereas collapse is a failure mode where the judge loses discriminative power or learns exploitable shortcuts.

\paragraph{Measuring Semantic Drift.}
We propose measuring semantic drift by tracking the consistency of rubric application across training iterations. Formally, given a rubric criterion $c_k$ and a fixed set of reference responses, we define the drift metric as:
\begin{equation}
\label{eq:semantic-drift}
\text{Drift}(c_k, t) = 1 - \text{corr}\big(s_t(c_k, x, y), s_0(c_k, x, y)\big)
\end{equation}
where $s_t$ is the criterion score at iteration $t$ and $\text{corr}$ measures correlation (e.g., Spearman's $\rho$). In practice, the reference set should be a frozen, diverse validation set of prompt--response pairs. Scores can be anchored to a static high-capacity judge or held-out human annotations to reduce model-dependent self-evaluation bias. Rank-correlation decay is then monitored across iterations, with intervention thresholds calibrated as an absolute or relative drop from the initial held-out correlation rather than a universal constant; crossing a threshold triggers rubric regeneration or human auditing. This diagnostic proposal is related to but distinct from semantic drift scores in text generation~\cite{knowwhen2024naacl} and value drift in post-training alignment~\cite{valuedrift2025}; its empirical validation remains open.

\subsection{Linguistic Reward Hacking}

\textit{Linguistic reward hacking} occurs when a policy exploits semantic gaps in natural language criteria, producing responses that satisfy the literal wording of a rubric while violating its intended meaning. This is a distinct phenomenon from numerical reward hacking (manipulating scalar reward magnitudes) and requires qualitative linguistic analysis to detect. While historically studied narrowly within low-resource language alignment~\cite{inuwadutse2025}, we generalize the concept to encompass any systematic policy exploitation of semantic gaps in multi-dimensional evaluation rubrics.

\paragraph{Root Causes: Ambiguity and Underspecification.}
This vulnerability stems directly from the fact that natural language criteria are inherently ambiguous and underspecified. Ambiguity arises because high-level desiderata like ``helpfulness'' depend heavily on latent user context, leading to only moderate alignment with human consensus (Kendall's $\tau$ of 0.3--0.4) even when mitigated by performance-level descriptors~\cite{hashemi2024llmrubric}. Underspecification occurs because criteria typically define \textit{what} to evaluate rather than \textit{how}, leaving operational boundaries vague unless anchored by intermediate granular structures like checklists~\cite{rlcf2025}. Crucially, ambiguity and underspecification function as the primary systemic drivers for both semantic drift and linguistic reward hacking (see detailed case studies in Appendix~\ref{sec:linguistic-appendix}). However, this underspecification is also an opportunity: it provides the essential semantic flexibility required for self-evolution and contextual adaptation. Without initial underspecification, optimization frameworks would lack the headroom to perform posterior sharpening or track emergent data distributions during training via automated alignment updates~\cite{evolm2026, rubricarm2026}.

\paragraph{Types of Linguistic Reward Hacking.}
We categorize these exploitative behaviors into three primary structural modes: (1) \textbf{Verbosity exploitation}, where the policy overoptimizes for ``completeness'' by generating elongated, low-substance responses, reflecting the proxy-reward degeneration patterns documented by \citet{gao2023scaling}. (2) \textbf{Criterion bleeding}, where the model satisfies one axis by sacrificing another, such as aggressively refusing valid prompts to maximize ``safety'' at the cost of ``helpfulness''~\cite{safeRLHF2024, armorm2024}. (3) \textbf{Surface pattern matching}, where the policy learns to parrot rubric keywords (e.g., stating ``I have ensured safety'') without executing the underlying reasoning. Self-evolving rubric work suggests that evaluator capacity and rubric specificity can affect how easily such surface-level satisfaction is detected~\cite{evolm2026}.

Potential diagnostics of linguistic reward hacking include: divergence between rubric scores and human preferences, auxiliary verification of claimed reasoning steps, abnormal criterion-correlation patterns, and cross-rubric consistency checks.

\paragraph{Relation to Bayesian Framework.}
Under the Bayesian lens (\S\ref{sec:framework}), linguistic reward hacking can be viewed as the policy exploiting the gap between a rubric's literal point-estimate instantiation ($\mathcal{R}_x$) and its intended conditional criterion space $P(\mathcal{R}|x)$. Optimizing across multiple plausible rubric formulations therefore suggests a robustness direction. Self-evolving systems may patch vulnerabilities as they are uncovered, but constitute approximate posterior updates only when the required probabilistic components are explicit.

\subsection{Cross-Lingual Considerations}

While rubric-guided RL is predominantly optimized for English, multilingual alignment introduces typological and systemic challenges. High-level desiderata such as ``respectfulness'' are culturally and linguistically mediated; Japanese and Korean honorific registers (\textit{keigo/jondaetmal}) and Arabic diglossia show that rubric and evaluator invariance cannot be assumed. A criterion may depend on dialect, speaker relation, register, and sociocultural setting rather than surface form alone. Low LLM-as-a-judge cross-lingual consistency ($\kappa \approx 0.3$) compounds this problem~\cite{fu2025multilingual}.

To mitigate these gaps, recent work on \textbf{cross-lingual reward transfer}
shows that English-trained reward models can bootstrap target-language
evaluations~\cite{hong2025crosslingual, implicitcrosslingual2025}, motivating
unified criteria sets and multilingual reward reasoning architectures
~\cite{sheth2026crosslingual, mr3_2026, mrewardbench2025}. We propose
\textit{language-aware rubric design}: decoupling language-agnostic properties
(e.g., factual accuracy) from language-specific instantiations (e.g., localized
formality). Under our Bayesian framework, this corresponds to conditioning on
language, dialect, speaker relation, and sociocultural context. For example,
$P(\mathcal{R}\mid x,\ell,d,z)$---while anchoring broadly shared principles in
the prior $P(\mathcal{R})$.

\subsection{Evidence Levels}

We distinguish three evidence levels of the analyses in this section: prompt sensitivity, evaluator calibration, granularity effects, and Goodhart-style over-optimization are established findings; semantic drift and linguistic reward hacking are our literature-grounded synthesis, supported by Appendix~\ref{sec:linguistic-appendix}; underspecification-as-opportunity and posterior marginalization are open hypotheses or design directions requiring validation.

%%%%%%%%%%%%%%
\section{Outlook and Emerging Frontiers: Agentic and Multimodal RL}
\label{sec:emerging-domains}

While our core taxonomy centers on text-based alignment, extending rubric-guided methods to agentic RL and multimodal models introduces fundamentally new structural challenges. This section highlights these emerging frontiers, which push rubric evaluation beyond single-turn chat, as summarized in the lower sections of Table~\ref{tab:taxonomy-v2}. Additional discussions on deep-research benchmarks are provided in Appendix~\ref{sec:app-agentic}.

\paragraph{Agentic RL and Credit Assignment.}
In agentic settings, evaluation must account for multi-step trajectories and environment grounding. The core bottleneck is \textit{multi-step credit assignment}. Traditional scalar rewards fail to isolate which specific action in a long trajectory led to success or failure. Rubric-guided methods like DR Tulu~\cite{drtulu2025}, HiPER~\cite{hiper2026}, and RubricEM~\cite{rubricem2026} address this by decomposing trajectory rewards into stagewise or hierarchical rubrics, enabling precise credit assignment across temporal phases. This structured feedback may reduce the policy from learning spurious correlations during complex tasks.

\paragraph{Multimodal Rubric-guided RL.}
In multimodal settings, evaluating outputs requires capturing cross-modal interactions and modality-specific dimensions such as visual grounding, object hallucination, and audio fluency. Foundational multimodal alignment work demonstrated that a single latent preference score is insufficient, by augmenting rewards with visual evidence~\cite{llavarlhf2024}, using segment-level correctional feedback~\cite{rlhfv2024}, separating safety and helpfulness objectives~\cite{saferlhfv2025}, or curating fine-grained multidimensional preference data~\cite{mmrlhf2025}. Recent methods directly address this complexity through explicit rubrics. Building on multidimensional benchmarks~\cite{pope2023, hallusionbench2024, mme2024}, systems such as Auto-Rubric as Reward~\cite{arr2026}, DeltaRubric~\cite{deltarubric2026}, and rDPO~\cite{rdpo2026} now explicitly condition multimodal reward modeling and preference optimization on structured criteria.
In the audio domain, Step-Audio-R1.5~\cite{stepaudior152026} applies rubric-guided optimization to multi-turn spoken dialogues by utilizing structural rubrics such as content requirements, persona consistency, and cross-turn instruction retention, while leaving subjective traits like tone and naturalness to standard preference comparison.

%%%%%%%%%%%%%%
\section{Conclusion and Future Directions}
\label{sec:conclusion}

Rubric-guided reinforcement learning marks a shift from implicit scalar reward
optimization to explicit, structured, and interpretable alignment. We unify this
space through a Bayesian framework that treats constitutions as priors
$P(\mathcal{R})$ and rubrics as conditional instantiations $P(\mathcal{R}|x)$,
interpreting some self-evolution mechanisms as approximate posterior updating
and rubric exploitation as point-estimate overfitting. Because rubrics are natural-language artifacts,
navigating granularity trade-offs and mitigating vulnerabilities like semantic drift and
linguistic reward hacking are central to building controllable, self-improving,
and reliable language models.

Future work faces several challenges. First, we need theory connecting
natural-language criteria to stable scalar or process-level feedback,
especially as rubrics and evaluators evolve.
Second, instance-specific rubric generation incurs token and latency overhead, motivating retrieval, caching, compression, and local evaluators.
Third, cross-lingual and multimodal frontiers require principled mappings from
abstract criteria to domain-specific instantiations. Finally, extending to
autonomous agents demands proper credit assignment over agentic tasks to reliably evaluate long-horizon trajectories.

%%%%%%%%%%%%%%
\section{Limitations}
\label{sec:limitations}

We acknowledge several limitations of this survey and of the rubric-guided RL paradigm more broadly.

\subsection{Heterogeneous Evaluation Settings}
Because rubric-guided RL methods use different base models, benchmarks, and
optimization protocols, direct quantitative comparison is difficult. We therefore
focus on qualitative organization and report numbers from original papers where
available. A standardized benchmark for controlled comparison remains an
important direction for future work.

\subsection{Scope of the Bayesian Framework}
Our Bayesian framework is an analytical abstraction for organizing how methods
specify, condition, and update evaluation criteria. It intentionally abstracts
over implementation details: fixed constitutions, deterministic rubric generators,
and self-evolving systems instantiate different parts of the notation. Exact
Bayesian guarantees would require explicit priors, likelihoods, observations, and
calibrated uncertainty, which most surveyed systems do not provide. Developing
fully probabilistic rubric generators remains an open research opportunity.

\subsection{Production Systems}
Recent technical reports suggest that production systems increasingly embed
rubric-like criteria in training and data pipelines, including Kimi K2,
GLM-5, Qwen3, and DeepSeek-V4~\cite{kimik22025, glm52026, qwen32025,
deepseekv42026}. For example, DeepSeek-V4 reports rubric-guided RL data and a
Generative Reward Model for hard-to-verify tasks, while GLM-5 defines
multidimensional criteria for chat and role-playing data, including instruction
following, expressiveness, creativity, coherence, and long-dialogue consistency
~\cite{deepseekv42026, glm52026}. Because these practices are often only
partially documented, a fuller account of implicit production-oriented rubric
usage remains an important direction for future empirical study.

\subsection{Evolving Terminology}
Terminology around constitutions, rubrics, and principles is still evolving.
Our framework provides one coherent vocabulary, but future work may refine these
distinctions as the field matures.

\subsection{Computational Cost}
As analyzed in \S\ref{sec:efficiency}, instance-specific rubric generation and API-based judging introduce token,
latency, and financial overhead. This creates deployment tradeoffs and motivates
cheaper rubric generation, caching, compression, and local evaluator designs.

%%%%%%%%%%%%%%
\section*{Ethics Statement}
\label{sec:ethics}
Rubric-guided RL raises important ethical considerations. While rubrics improve interpretability, they also pose the systemic risk of embedding cultural, political, or institutional biases into formalized evaluation criteria that are subsequently treated as objective standards. Evaluator dependence, underspecified criteria, and aggregation choices can further conceal whose values dominate. The sensitivity to normative framing observed in Constitutional AI~\cite{bai2022constitutional} extends to rubrics more broadly. Future work should develop frameworks for ethical rubric design by prioritizing participatory construction to engage diverse stakeholders in defining criteria, democratic governance to establish formal mechanisms for continuous revision and contestation, and systemic transparency to ensure that both the underlying criteria and their downstream policy updates remain fully auditable.

\section*{Acknowledgments}
\label{sec:ack}

AI assistants, including ChatGPT and Gemini, were used only to identify additional relevant work and to polish author-written text. All suggested references and language edits were independently verified by the authors.

We thank Yao Zhou and Zhenran Xu for their valuable discussions.

\bibliography{references}

@inproceedings{ouyang2022training,
  title = {Training language models to follow instructions with human feedback},
  author = {Ouyang, Long and Wu, Jeffrey and Jiang, Xu and Almeida, Diogo and Wainwright, Carroll and Mishkin, Pamela and Zhang, Chong and Agarwal, Sandhini and Slama, Katarina and Ray, Alex and Schulman, John and Hilton, Jacob and Kelton, Fraser and Miller, Luke and Simens, Max and Askell, Amanda and Welinder, Peter and Christiano, Paul and Leike, Jan and Lowe, Ryan},
  booktitle = {Advances in Neural Information Processing Systems},
  volume = {35},
  pages = {27730--27744},
  year = {2022},
  doi = {10.52202/068431-2011},
  url = {https://doi.org/10.52202/068431-2011}
}

@article{bai2022constitutional,
  title = {Constitutional {AI}: Harmlessness from {AI} feedback},
  author = {Yuntao Bai and Saurav Kadavath and Sandipan Kundu and Amanda Askell and Jackson Kernion and Andy Jones and Anna Chen and Anna Goldie and Azalia Mirhoseini and Cameron McKinnon and Carol Chen and Catherine Olsson and Christopher Olah and Danny Hernandez and Dawn Drain and Deep Ganguli and Dustin Li and Eli Tran-Johnson and Ethan Perez and Jamie Kerr and Jared Mueller and Jeffrey Ladish and Joshua Landau and Kamal Ndousse and Kamile Lukosuite and Liane Lovitt and Michael Sellitto and Nelson Elhage and Nicholas Schiefer and Noemi Mercado and Nova DasSarma and Robert Lasenby and Robin Larson and Sam Ringer and Scott Johnston and Shauna Kravec and Sheer El Showk and Stanislav Fort and Tamera Lanham and Timothy Telleen-Lawton and Tom Conerly and Tom Henighan and Tristan Hume and Samuel R. Bowman and Zac Hatfield-Dodds and Ben Mann and Dario Amodei and Nicholas Joseph and Sam McCandlish and Tom Brown and Jared Kaplan},
  journal = {arXiv preprint arXiv:2212.08073},
  year = {2022},
  url = {https://arxiv.org/abs/2212.08073}
}

@inproceedings{rafailov2023dpo,
  title = {Direct preference optimization: Your language model is secretly a reward model},
  author = {Rafailov, Rafael and Sharma, Archit and Mitchell, Eric and Manning, Christopher D. and Ermon, Stefano and Finn, Chelsea},
  booktitle = {Advances in Neural Information Processing Systems},
  volume = {36},
  pages = {53728-53741},
  year = {2023},
  doi = {10.52202/075280-2338},
  url = {https://doi.org/10.52202/075280-2338}
}

@inproceedings{lightman2023lets,
  title = {Let's verify step by step},
  author = {Lightman, Hunter and Kosaraju, Vineet and Burda, Yuri and Edwards, Harri and Baker, Bowen and Lee, Teddy and Leike, Jan and Schulman, John and Sutskever, Ilya and Cobbe, Karl},
  booktitle = {International Conference on Learning Representations},
  year = {2024},
  url = {https://arxiv.org/abs/2305.20050}
}

@inproceedings{lee2023rlaif,
  title = {{RLAIF} vs. {RLHF}: Scaling Reinforcement Learning from Human Feedback with {AI} Feedback},
  author = {Lee, Harrison and Phatale, Samrat and Mansoor, Hassan and Mesnard, Thomas and Ferret, Johan and Lu, Kellie and Bishop, Colton and Hall, Ethan and Carbune, Victor and Rastogi, Abhinav and Prakash, Sushant},
  booktitle = {International Conference on Machine Learning},
  series = {Proceedings of Machine Learning Research},
  volume = {235},
  pages = {26874--26901},
  year = {2024},
  url = {https://arxiv.org/abs/2309.00267}
}

@article{shao2024deepseekmath,
  title = {{DeepSeekMath}: Pushing the Limits of Mathematical Reasoning in Open Language Models},
  author = {Shao, Zhihong and Wang, Peiyi and Zhu, Qihao and Xu, Runxin and Song, Junxiao and Bi, Xiao and Zhang, Haowei and Zhang, Mingchuan and Li, Y. K. and Wu, Y. and Guo, Daya},
  journal = {arXiv preprint arXiv:2402.03300},
  year = {2024},
  url = {https://arxiv.org/abs/2402.03300}
}

@inproceedings{rar2025,
  title = {Rubrics as Rewards: Reinforcement Learning Beyond Verifiable Domains},
  author = {Anisha Gunjal and Anthony Wang and Elaine Lau and Vaskar Nath and Yunzhong He and Bing Liu and Sean M. Hendryx},
  booktitle = {The Fourteenth International Conference on Learning Representations},
  year = {2026},
  url = {https://openreview.net/forum?id=c1bTcrDmt4}
}

@article{rubricarm2026,
  title = {Alternating Reinforcement Learning for Rubric-Based Reward Modeling in Non-Verifiable {LLM} Post-Training},
  author = {Xu, Ran and Liu, Tianci and Dong, Zihan and Yu, Tony and Hong, Ilgee and Yang, Carl and Zhang, Linjun and Zhao, Tao and Wang, Haoyu},
  journal = {arXiv preprint arXiv:2602.01511},
  year = {2026},
  url = {https://arxiv.org/abs/2602.01511}
}

@inproceedings{evolm2026,
  title = {{EvoLM}: Self-Evolving Language Models through Co-Evolved Discriminative Rubrics},
  author = {Shuyue Stella Li and Rui Xin and Teng Xiao and Yike Wang and Rulin Shao and Zoey Hao and Melanie Sclar and Sewoong Oh and Faeze Brahman and Pang Wei Koh and Yulia Tsvetkov},
  booktitle = {The Fourteenth International Conference on Learning Representations},
  year = {2026},
  url = {https://openreview.net/forum?id=aA2PXFH2Cp}
}

@article{rlcer2026,
  title = {Reinforcing Chain-of-Thought Reasoning with Self-Evolving Rubrics},
  author = {Sheng, Leheng and Ma, Wenchang and Hong, Ruixin and Wang, Xiang and Zhang, An and Chua, Tat-Seng},
  journal = {arXiv preprint arXiv:2602.10885},
  year = {2026},
  url = {https://arxiv.org/abs/2602.10885}
}

@inproceedings{openrubrics2025,
  title = {{O}pen{R}ubrics: Towards Scalable Synthetic Rubric Generation for Reward Modeling and {LLM} Alignment},
  author = {Liu, Tianci and Xu, Ran and Yu, Tony and Hong, Ilgee and Yang, Carl and Zhao, Tuo and Wang, Haoyu},
  editor = {Liakata, Maria and Moreira, Viviane P. and Zhang, Jiajun and Jurgens, David},
  booktitle = {Proceedings of the 64th Annual Meeting of the {A}ssociation for {C}omputational {L}inguistics (Volume 1: Long Papers)},
  pages = {17417--17437},
  year = {2026},
  address = {San Diego, California, United States},
  publisher = {Association for Computational Linguistics},
  doi = {10.18653/v1/2026.acl-long.791},
  url = {https://aclanthology.org/2026.acl-long.791/},
  isbn = {979-8-89176-390-6}
}

@inproceedings{hashemi2024llmrubric,
  title = {{LLM}-Rubric: A Multidimensional, Calibrated Approach to Automated Evaluation of Natural Language Texts},
  author = {Hashemi, Helia and Eisner, Jason and Rosset, Corby and Van Durme, Benjamin and Kedzie, Chris},
  editor = {Ku, Lun-Wei and Martins, Andre and Srikumar, Vivek},
  booktitle = {Proceedings of the 62nd Annual Meeting of the Association for Computational Linguistics (Volume 1: Long Papers)},
  pages = {13806--13834},
  year = {2024},
  address = {Bangkok, Thailand},
  publisher = {Association for Computational Linguistics},
  doi = {10.18653/v1/2024.acl-long.745},
  url = {https://aclanthology.org/2024.acl-long.745/}
}

@article{gu2024survey,
  title = {A Survey on {LLM-as-a-Judge}},
  author = {Gu, Jiawei and Jiang, Xuhui and Shi, Zhichao and Tan, Hexiang and Zhai, Xuehao and Xu, Chengjin and Li, Wei and Shen, Yinghan and Ma, Shengjie and Liu, Honghao and Wang, Saizhuo and Zhang, Kun and Wang, Yuanzhuo and Gao, Wen and Ni, Lionel and Guo, Jian},
  journal = {arXiv preprint arXiv:2411.15594},
  year = {2024},
  url = {https://arxiv.org/abs/2411.15594}
}

@inproceedings{iteralign2024,
  title = {{I}ter{A}lign: Iterative Constitutional Alignment of Large Language Models},
  author = {Chen, Xiusi and Wen, Hongzhi and Nag, Sreyashi and Luo, Chen and Yin, Qingyu and Li, Ruirui and Li, Zheng and Wang, Wei},
  editor = {Duh, Kevin and Gomez, Helena and Bethard, Steven},
  booktitle = {Proceedings of the 2024 Conference of the North American Chapter of the Association for Computational Linguistics: Human Language Technologies (Volume 1: Long Papers)},
  pages = {1423--1433},
  year = {2024},
  address = {Mexico City, Mexico},
  publisher = {Association for Computational Linguistics},
  doi = {10.18653/v1/2024.naacl-long.78},
  url = {https://aclanthology.org/2024.naacl-long.78/}
}

@inproceedings{icai2024,
  title = {Inverse Constitutional {AI}: Compressing Preferences into Principles},
  author = {Findeis, Arduin and Kaufmann, Timo and H{\"u}llermeier, Eyke and Albanie, Samuel and Mullins, Robert},
  booktitle = {International Conference on Learning Representations},
  year = {2025},
  url = {https://arxiv.org/abs/2406.06560}
}

@article{deliberative2024,
  title = {Deliberative Alignment: Reasoning Enables Safer Language Models},
  author = {OpenAI},
  journal = {arXiv preprint arXiv:2412.16339},
  year = {2024},
  url = {https://arxiv.org/abs/2412.16339}
}

@inproceedings{finegrained2023,
  title = {Fine-Grained Human Feedback Gives Better Rewards for Language Model Training},
  author = {Wu, Zeqiu and Hu, Yushi and Shi, Weijia and Dziri, Nouha and Suhr, Alane and Ammanabrolu, Prithviraj and Smith, Noah A. and Ostendorf, Mari and Hajishirzi, Hannaneh},
  booktitle = {Advances in Neural Information Processing Systems},
  volume = {36},
  pages = {59008-59033},
  year = {2023},
  doi = {10.52202/075280-2574},
  url = {https://doi.org/10.52202/075280-2574}
}

@article{rubricbench2026,
  title = {{RubricBench}: Aligning Model-Generated Rubrics with Human Standards},
  author = {Zhang, Qiyuan and Zhou, Junyi and Wang, Yufei and Lyu, Fuyuan and Ming, Yidong and Xu, Can and Sun, Qingfeng and Zheng, Kai and Kang, Peng and Liu, Xue and Ma, Chen},
  journal = {arXiv preprint arXiv:2603.01562},
  year = {2026},
  url = {https://arxiv.org/abs/2603.01562}
}

@article{rubriceval2026,
  title = {{RubricEval}: A Rubric-Level Meta-Evaluation Benchmark for {LLM} Judges in Instruction Following},
  author = {Pan, Tianjun and Lin, Xuan and Yang, Wenyan and He, Qianyu and Chen, Shisong and Qi, Licai and Xu, Wanqing and Feng, Hongwei and Xu, Bo and Xiao, Yanghua},
  journal = {arXiv preprint arXiv:2603.25133},
  year = {2026},
  url = {https://arxiv.org/abs/2603.25133}
}

@inproceedings{chasingtail2025,
  title = {Chasing the Tail: Effective Rubric-Based Reward Modeling for Large Language Model Post-Training},
  author = {Junkai Zhang and Zihao Wang and Lin Gui and Swarnashree Mysore Sathyendra and Jaehwan Jeong and Victor Veitch and Wei Wang and Yunzhong He and Bing Liu and Lifeng Jin},
  booktitle = {The Fourteenth International Conference on Learning Representations},
  year = {2026},
  url = {https://openreview.net/forum?id=pBjy4ek2QV}
}

@inproceedings{critiquellm2024,
  title = {{C}ritique{LLM}: Towards an Informative Critique Generation Model for Evaluation of Large Language Model Generation},
  author = {Ke, Pei and Wen, Bosi and Feng, Andrew and Liu, Xiao and Lei, Xuanyu and Cheng, Jiale and Wang, Shengyuan and Zeng, Aohan and Dong, Yuxiao and Wang, Hongning and Tang, Jie and Huang, Minlie},
  editor = {Ku, Lun-Wei and Martins, Andre and Srikumar, Vivek},
  booktitle = {Proceedings of the 62nd Annual Meeting of the Association for Computational Linguistics (Volume 1: Long Papers)},
  pages = {13034--13054},
  year = {2024},
  address = {Bangkok, Thailand},
  publisher = {Association for Computational Linguistics},
  doi = {10.18653/v1/2024.acl-long.704},
  url = {https://aclanthology.org/2024.acl-long.704/}
}

@article{tao2024selfevolution,
  title = {A Survey on Self-Evolution of Large Language Models},
  author = {Tao, Zhengwei and Lin, Ting-En and Sun, Qinyuan and Li, Xingjian and Zhang, Hongling and Zhang, Zhe and Lian, Yuting and Sun, Bei and Wang, Qifan and Li, Junwei and Wu, Zhiqiang and Zhou, Jingren},
  journal = {arXiv preprint arXiv:2404.14387},
  year = {2024},
  url = {https://arxiv.org/abs/2404.14387}
}

@inproceedings{safeRLHF2024,
  title = {Safe {RLHF}: Safe Reinforcement Learning from Human Feedback},
  author = {Dai, Josef and Pan, Rui and Ji, Jiaming and Li, Xinyu and Wang, Yaodong},
  booktitle = {International Conference on Learning Representations},
  year = {2024},
  url = {https://arxiv.org/abs/2310.12773}
}

@inproceedings{saferlhfv2025,
  title = {Safe {RLHF-V}: Safe Reinforcement Learning from Multi-modal Human Feedback},
  author = {Ji, Jiaming and Chen, Xinyu and Pan, Rui and Zhu, Han and Li, Jiahao and Hong, Donghai and Chen, Boyuan and Zhou, Jiayi and Wang, Kaile and Dai, Juntao and Chan, Chi-Min and Tang, Yida and Yang, Yaodong},
  booktitle = {Advances in Neural Information Processing Systems},
  pages = {51568-51604},
  year = {2025},
  doi = {10.52202/085713-1537},
  url = {https://doi.org/10.52202/085713-1537}
}

@inproceedings{inform2024,
  title = {{InfoRM}: Mitigating Reward Hacking in {RLHF} via Information-Theoretic Reward Modeling},
  author = {Miao, Yuchun and Zhang, Sen and Ding, Liang and Bao, Rong and Zhang, Lefei and Tao, Dacheng},
  booktitle = {Advances in Neural Information Processing Systems},
  pages = {134387-134429},
  year = {2024},
  doi = {10.52202/079017-4270},
  url = {https://doi.org/10.52202/079017-4270}
}

@inproceedings{ethayarajh2024kto,
  title = {{KTO}: Model Alignment as Prospect Theoretic Optimization},
  author = {Ethayarajh, Kawin and Xu, Winnie and Muennighoff, Niklas and Jurafsky, Dan and Kiela, Douwe},
  booktitle = {International Conference on Machine Learning},
  year = {2024},
  url = {https://arxiv.org/abs/2402.01306}
}

@article{rewarddesign2025,
  title = {A Survey on Progress in {LLM} Alignment from the Perspective of Reward Design},
  author = {Ji, Miaomiao and Wu, Yanqiu and Wu, Zhibin and Wang, Shoujin and Yang, Jian and Dras, Mark and Naseem, Usman},
  journal = {arXiv preprint arXiv:2505.02666},
  year = {2025},
  url = {https://arxiv.org/abs/2505.02666}
}

@article{prmsurvey2025,
  title = {A Survey of Process Reward Models: From Outcome Signals to Process Supervisions for Large Language Models},
  author = {Zheng, Congmin and Zhu, Jiachen and Ou, Zhuoying and Chen, Yuxiang and Zhang, Kangning and Shan, Rong and Zheng, Zeyu and Yang, Mengyue and Lin, Jianghao and Yu, Yong and Zhang, Weinan},
  journal = {arXiv preprint arXiv:2510.08049},
  year = {2025},
  url = {https://arxiv.org/abs/2510.08049}
}

@article{autorubric2026,
  title = {{Autorubric}: A Unifying Framework for Rubric-Based {LLM} Evaluation on Non-Verifiable Tasks},
  author = {Rao, Delip and Callison-Burch, Chris},
  journal = {arXiv preprint arXiv:2603.00077},
  year = {2026},
  url = {https://arxiv.org/abs/2603.00077}
}

@inproceedings{learntojudge2026,
  title = {Learning to Judge: {LLM}s Designing and Applying Evaluation Rubrics},
  author = {Siro, Clemencia and Aliannejadi, Pourya and Aliannejadi, Mohammad},
  editor = {Demberg, Vera and Inui, Kentaro and Marquez, Llu{\'i}s},
  booktitle = {Findings of the {A}ssociation for {C}omputational {L}inguistics: {EACL} 2026},
  pages = {6371--6389},
  year = {2026},
  address = {Rabat, Morocco},
  publisher = {Association for Computational Linguistics},
  doi = {10.18653/v1/2026.findings-eacl.335},
  url = {https://aclanthology.org/2026.findings-eacl.335/},
  isbn = {979-8-89176-386-9}
}

@article{arr2026,
  title = {Auto-Rubric as Reward: From Implicit Preferences to Explicit Multimodal Generative Criteria},
  author = {Tian, Juanxi and Liu, Fengyuan and Han, Jiaming and Jiang, Yilei and Wu, Yongliang and Liu, Yesheng and Li, Haodong and Xu, Furong and Li, Wanhua},
  journal = {arXiv preprint arXiv:2605.08354},
  year = {2026},
  url = {https://arxiv.org/abs/2605.08354}
}

@article{deltarubric2026,
  title = {{DeltaRubric}: Generative Multimodal Reward Modeling via Joint Planning and Verification},
  author = {Liu, Rui and Yu, Dian and Liang, Zhenwen and Shi, Yucheng and Zheng, Tong and Dai, Runpeng and Mi, Haitao and Tokekar, Pratap and Leoweiliang},
  journal = {arXiv preprint arXiv:2605.09269},
  year = {2026},
  url = {https://arxiv.org/abs/2605.09269}
}

@article{rdpo2026,
  title = {Visual Preference Optimization with Rubric Rewards},
  author = {Yu, Ya-Qi and Hong, Fangyu and Qu, Xiangyang and Wang, Hao and Wu, Gaojie and Luo, Qiaoyu and Xu, Nuo and Wang, Huixin and Xu, Wuheng and Liao, Yongxin and Chen, Zihao and Li, Haonan and Li, Ziming and Peng, Dezhi and Liao, Minghui and Wu, Jihao and Ren, Haoyu and Tu, Dandan},
  journal = {arXiv preprint arXiv:2604.13029},
  year = {2026},
  url = {https://arxiv.org/abs/2604.13029}
}

@inproceedings{mmrlhf2025,
  title = {{MM-RLHF}: The Next Step Forward in Multimodal {LLM} Alignment},
  author = {Zhang, Yi-Fan and Yu, Tao and Tian, Haochen and Fu, Chaoyou and Li, Peiyan and Zeng, Jianshu and Xie, Wulin and Shi, Yang and Zhang, Huanyu and Wang, Xue and Hu, Yibo and Luo, Bin and Tan, Tien-Ping},
  booktitle = {International Conference on Machine Learning},
  year = {2025},
  url = {https://arxiv.org/abs/2502.10391}
}

@article{rubrichacking2026,
  title = {Reward Hacking in Rubric-Based Reinforcement Learning},
  author = {Mahmoud, Anas and Rezaei, MohammadHossein and Wang, Zihao and Gunjal, Anisha and Liu, Bing and He, Yunzhong},
  journal = {arXiv preprint arXiv:2605.12474},
  year = {2026},
  url = {https://arxiv.org/abs/2605.12474}
}

@inproceedings{drtulu2025,
  title = {{{DR Tulu}: Reinforcement Learning with Evolving Rubrics for Deep Research}},
  author = {Rulin Shao and Akari Asai and Shannon Zejiang Shen and Hamish Ivison and Varsha Kishore and Jingming Zhuo and Xinran Zhao and Molly Park and Samuel G. Finlayson and David Sontag and Tyler Murray and Sewon Min and Pradeep Dasigi and Luca Soldaini and Faeze Brahman and Wen-tau Yih and Tongshuang Wu and Luke Zettlemoyer and Yoon Kim and Hannaneh Hajishirzi and Pang Wei Koh},
  booktitle = {International Conference on Machine Learning},
  year = {2026},
  url = {https://arxiv.org/abs/2511.19399}
}

@article{adarubric2026,
  title = {{AdaRubric}: Task-Adaptive Rubrics for Reliable LLM Agent Evaluation and Reward Learning},
  author = {Ding, Liang},
  journal = {arXiv preprint arXiv:2603.21362},
  year = {2026},
  url = {https://arxiv.org/abs/2603.21362},
  note = {KnowFM Workshop at ACL 2026}
}

@article{openrs2026,
  title = {Open Rubric System: Scaling Reinforcement Learning with Pairwise Adaptive Rubric},
  author = {Jia, Ruipeng and Yang, Yunyi and Wu, Yuxin and Gai, Yongbo and Tao, Siyuan and Zhou, Mengyu and Lin, Jianhe and Jiang, Xiaoxi and Jiang, Guanjun},
  journal = {arXiv preprint arXiv:2602.14069},
  year = {2026},
  url = {https://arxiv.org/abs/2602.14069}
}

@article{hiper2026,
  title = {{HiPER}: Hierarchical Reinforcement Learning with Explicit Credit Assignment for Large Language Model Agents},
  author = {Peng, Jiangweizhi and Liu, Yuanxin and Zhou, Ruida and Fleming, Charles and Wang, Zhaoran and Garcia, Alfredo and Hong, Mingyi},
  journal = {arXiv preprint arXiv:2602.16165},
  year = {2026},
  url = {https://arxiv.org/abs/2602.16165}
}

@inproceedings{rlcf2025,
  title = {Checklists Are Better Than Reward Models For Aligning Language Models},
  author = {Viswanathan, Vijay and Sun, Yanchao and Ma, Shuang and Kong, Xiang and Cao, Meng and Neubig, Graham and Wu, Tongshuang},
  booktitle = {Advances in Neural Information Processing Systems},
  year = {2025},
  url = {https://arxiv.org/abs/2507.18624}
}

@misc{kimik22025,
  title = {{Kimi K2}: Open Agentic Intelligence},
  author = {{Kimi Team} and Yifan Bai and Yiping Bao and Guanduo Chen and Jiahao Chen and Ningxin Chen and Ruijue Chen and Yanru Chen and Yuankun Chen and Yutian Chen and others},
  year = {2025},
  url = {https://arxiv.org/abs/2507.20534},
  eprint = {2507.20534},
  archiveprefix = {arXiv},
  primaryclass = {cs.CL}
}

@article{glm52026,
  title = {{GLM-5}: From Vibe Coding to Agentic Engineering},
  author = {{GLM-5 Team}},
  journal = {arXiv preprint arXiv:2602.15763},
  year = {2026},
  url = {https://arxiv.org/abs/2602.15763}
}

@article{qwen32025,
  title = {{Qwen3} Technical Report},
  author = {An Yang and Anfeng Li and Baosong Yang and Beichen Zhang and Binyuan Hui and Bo Zheng and Bowen Yu and Chang Gao and Chengen Huang and Chenxu Lv and Chujie Zheng and Dayiheng Liu and Fan Zhou and Fei Huang and Feng Hu and Hao Ge and Haoran Wei and Huan Lin and Jialong Tang and Jian Yang and Jianhong Tu and Jianwei Zhang and Jianxin Yang and Jiaxi Yang and Jing Zhou and Jingren Zhou and Junyang Lin and Kai Dang and Keqin Bao and Kexin Yang and Le Yu and Lianghao Deng and Mei Li and Mingfeng Xue and Mingze Li and Pei Zhang and Peng Wang and Qin Zhu and Rui Men and Ruize Gao and Shixuan Liu and Shuang Luo and Tianhao Li and Tianyi Tang and Wenbiao Yin and Xingzhang Ren and Xinyu Wang and Xinyu Zhang and Xuancheng Ren and Yang Fan and Yang Su and Yichang Zhang and Yinger Zhang and Yu Wan and Yuqiong Liu and Zekun Wang and Zeyu Cui and Zhenru Zhang and Zhipeng Zhou and Zihan Qiu},
  journal = {arXiv preprint arXiv:2505.09388},
  year = {2025},
  url = {https://arxiv.org/abs/2505.09388}
}

@techreport{deepseekv42026,
  title = {{DeepSeek-V4}: Towards Highly Efficient Million-Token Context Intelligence},
  author = {{DeepSeek-AI}},
  year = {2026},
  institution = {DeepSeek-AI},
  doi = {10.48550/arXiv.2606.19348},
  url = {https://arxiv.org/abs/2606.19348}
}

@article{clbench2026,
  title = {{CL-bench}: A Benchmark for Context Learning},
  author = {Dou, Shihan and Zhang, Ming and Yin, Zhangyue and Huang, Chenhao and Shen, Yujiong and Wang, Junzhe and Chen, Jiayi and Ni, Yuchen and Ye, Junjie and Zhang, Cheng and Xie, Huaibing and Hu, Jianglu and Wang, Shaolei and Wang, Weichao and Xiao, Yanling and Liu, Yiting and Xu, Zenan and Guo, Zhen and Zhou, Pluto and Gui, Tao and Wu, Zuxuan and Qiu, Xipeng and Zhang, Qi and Huang, Xuanjing and Jiang, Yu-Gang and Wang, Di and Yao, Shunyu},
  journal = {arXiv preprint arXiv:2602.03587},
  year = {2026},
  url = {https://arxiv.org/abs/2602.03587}
}

@article{clbenchlife2026,
  title = {{CL-bench Life}: Can Language Models Learn from Real-Life Context?},
  author = {Dou, Shihan and Shen, Yujiong and Huang, Chenhao and Ye, Junjie and Chen, Jiayi and Wang, Junzhe and He, Qianyu and Liu, Shichun and Lv, Changze and Lin, Jiahang and Zhang, Jiazheng and Zhang, Ming and Liu, Shaofan and Ji, Tao and Yin, Zhangyue and Zhang, Cheng and Xie, Huaibing and Hu, Jianglu and Deng, Jingcheng and Li, Lincheng and Hu, Minda and Wang, Shaolei and Zhao, Syrus and Wang, Weichao and Lei, Yan and Liu, Yang and Xiao, Yanling and Liu, Yiting and Xu, Zenan and Guo, Zhen and Zhao, Ziliang and Zhou, Pluto and Gui, Tao and Zhang, Qi and Huang, Xuanjing and Jiang, Yu-Gang and Wang, Di and Yao, Shunyu},
  journal = {arXiv preprint arXiv:2604.27043},
  year = {2026},
  url = {https://arxiv.org/abs/2604.27043}
}

@article{rubricem2026,
  title = {{RubricEM}: {Meta-RL} with Rubric-guided Policy Decomposition beyond Verifiable Rewards},
  author = {Li, Gaotang and Mishra, Bhavana Dalvi and Wang, Zifeng and Yan, Jun and Chen, Yanfei and Li, Chun-Liang and Le, Long T. and Han, Rujun and Lee, George and Tong, Hanghang and Lee, Chen-Yu and Pfister, Tomas},
  journal = {arXiv preprint arXiv:2605.10899},
  year = {2026},
  url = {https://arxiv.org/abs/2605.10899}
}

@inproceedings{qalign2025,
  title = {{QA}{-}{LIGN}: Aligning {LLM}s through Constitutionally Decomposed {QA}},
  author = {Dineen, Jacob and Rrv, Aswin and Liu, Qin and Xu, Zhikun and Ye, Xiao and Shen, Ming and Li, Zhaonan and Lu, Shijie and Baral, Chitta and Chen, Muhao and Zhou, Ben},
  editor = {Christodoulopoulos, Christos and Chakraborty, Tanmoy and Rose, Carolyn and Peng, Violet},
  booktitle = {Findings of the Association for Computational Linguistics: EMNLP 2025},
  pages = {20619--20642},
  year = {2025},
  address = {Suzhou, China},
  publisher = {Association for Computational Linguistics},
  doi = {10.18653/v1/2025.findings-emnlp.1123},
  url = {https://aclanthology.org/2025.findings-emnlp.1123/},
  isbn = {979-8-89176-335-7}
}

@article{rrd2026,
  title = {Rethinking Rubric Generation for Improving {LLM} Judge and Reward Modeling for Open-ended Tasks},
  author = {William F. Shen and Xinchi Qiu and Chenxi Whitehouse and Lisa Alazraki and Shashwat Goel and Francesco Barbieri and Timon Willi and Akhil Mathur and Ilias Leontiadis},
  journal = {arXiv preprint arXiv:2602.05125},
  year = {2026},
  url = {https://arxiv.org/abs/2602.05125}
}

@article{draco2026,
  title = {{DRACO}: A Cross-Domain Benchmark for Deep Research Accuracy, Completeness, and Objectivity},
  author = {Zhong, Joey and Zhang, Hao and Southern, Clare and Yang, Jeremy and Wang, Thomas and Jung, Kate and Zhang, Shu and Yarats, Denis and Ho, Johnny and Ma, Jerry},
  journal = {arXiv preprint arXiv:2602.11685},
  year = {2026},
  url = {https://arxiv.org/abs/2602.11685}
}

@article{researchrubrics2026,
  title = {{ResearchRubrics}: A Benchmark of Prompts and Rubrics For Evaluating Deep Research Agents},
  author = {Sharma, Manasi and Zhang, Chen Bo Calvin and Bandi, Chaithanya and Wang, Clinton and Aich, Ankit and Nghiem, Huy and Rabbani, Tahseen and Htet, Ye and Jang, Brian and Basu, Sumana and Balwani, Aishwarya and Peskoff, Denis and Ayestaran, Marcos and Hendryx, Sean M. and Kenstler, Brad and Liu, Bing},
  journal = {arXiv preprint arXiv:2511.07685},
  year = {2025},
  url = {https://arxiv.org/abs/2511.07685}
}

@article{deepresearchbench2026,
  title = {{DeepResearch Bench II}: Diagnosing Deep Research Agents via Rubrics from Expert Report},
  author = {Li, Ruizhe and Du, Mingxuan and Xu, Benfeng and Zhu, Chiwei and Wang, Xiaorui and Mao, Zhendong},
  journal = {arXiv preprint arXiv:2601.08536},
  year = {2026},
  url = {https://arxiv.org/abs/2601.08536}
}

@article{deer2025,
  title = {{DEER}: A Benchmark for Evaluating Deep Research Agents on Expert Report Generation},
  author = {Janghoon Han and Heegyu Kim and Changho Lee and Dahm Lee and Min Hyung Park and Hosung Song and Stanley Jungkyu Choi and Moontae Lee and Honglak Lee},
  journal = {arXiv preprint arXiv:2512.17776},
  year = {2025},
  url = {https://arxiv.org/abs/2512.17776}
}

@article{gama2014survey,
  author    = {Gama, Jo{\~{a}}o and {\v{Z}}liobait{\.{e}}, Indr{\.{e}} and Bifet, Albert and Pechenizkiy, Mykola and Bouchachia, Abdelhamid},
  title     = {A Survey on Concept Drift Adaptation},
  journal   = {ACM Computing Surveys},
  volume    = {46},
  number    = {4},
  pages     = {44:1--44:37},
  year      = {2014},
  publisher = {ACM},
  doi       = {10.1145/2523813},
  url={https://doi.org/10.1145/2523813}
}

@inproceedings{knowwhen2024naacl,
  title = {Know When To Stop: A Study of Semantic Drift in Text Generation},
  author = {Spataru, Ava and Hambro, Eric and Voita, Elena and Cancedda, Nicola},
  editor = {Duh, Kevin and Gomez, Helena and Bethard, Steven},
  booktitle = {Proceedings of the 2024 Conference of the North American Chapter of the Association for Computational Linguistics: Human Language Technologies (Volume 1: Long Papers)},
  pages = {3656--3671},
  year = {2024},
  address = {Mexico City, Mexico},
  publisher = {Association for Computational Linguistics},
  doi = {10.18653/v1/2024.naacl-long.202},
  url = {https://aclanthology.org/2024.naacl-long.202/}
}

@article{valuedrift2025,
  title = {Value Drifts: Tracing Value Alignment During {LLM} Post-Training},
  author = {Bhatia, Mehar and Nayak, Shravan and Kamath, Gaurav and Mosbach, Marius and Sta{\'n}czak, Karolina and Shwartz, Vered and Reddy, Siva},
  journal = {arXiv preprint arXiv:2510.26707},
  year = {2025},
  url = {https://arxiv.org/abs/2510.26707}
}

@article{inuwadutse2025,
  title = {{OpenAI}'s {GPT-OSS-20B} Model and Safety Alignment Issues in a Low-Resource Language},
  author = {Inuwa-Dutse, Isa},
  journal = {arXiv preprint arXiv:2510.01266},
  year = {2025},
  url = {https://arxiv.org/abs/2510.01266}
}

@inproceedings{armorm2024,
  title = {Interpretable Preferences via Multi-Objective Reward Modeling and Mixture-of-Experts},
  author = {Wang, Haoxiang and Xiong, Wei and Xie, Tengyang and Zhao, Han and Zhang, Tong},
  editor = {Al-Onaizan, Yaser and Bansal, Mohit and Chen, Yun-Nung},
  booktitle = {Findings of the Association for Computational Linguistics: EMNLP 2024},
  pages = {10582--10592},
  year = {2024},
  address = {Miami, Florida, USA},
  publisher = {Association for Computational Linguistics},
  doi = {10.18653/v1/2024.findings-emnlp.620},
  url = {https://aclanthology.org/2024.findings-emnlp.620/}
}

@inproceedings{hong2025crosslingual,
  title = {Cross-lingual Transfer of Reward Models in Multilingual Alignment},
  author = {Hong, Jiwoo and Lee, Noah and Mart{\'i}nez-Casta{\~n}o, Rodrigo and Rodr{\'i}guez, C{\'e}sar and Thorne, James},
  booktitle = {Proceedings of the 2025 Conference of the North American Chapter of the Association for Computational Linguistics},
  year = {2025},
  url = {https://arxiv.org/abs/2410.18027}
}

@article{sheth2026crosslingual,
  title = {Cross-Lingual {LLM-Judge} Transfer via Evaluation Decomposition},
  author = {Sheth, Ivaxi and Jonke, Zeno and Mantrach, Amin and Mansour, Saab},
  journal = {arXiv preprint arXiv:2603.18557},
  year = {2026},
  url = {https://arxiv.org/abs/2603.18557}
}

@inproceedings{mr3_2026,
  title = {{mR3}: Multilingual Rubric-Agnostic Reward Reasoning Models},
  author = {David Anugraha and Shou-Yi Hung and Zilu Tang and En-Shiun Annie Lee and Derry Tanti Wijaya and Genta Indra Winata},
  booktitle = {The Fourteenth International Conference on Learning Representations},
  year = {2026},
  url = {https://openreview.net/forum?id=ST0wOB1bdX}
}

@inproceedings{mrewardbench2025,
  title = {{M}-{R}eward{B}ench: Evaluating Reward Models in Multilingual Settings},
  author = {Gureja, Srishti and Miranda, Lester James V. and Islam, Shayekh Bin and Maheshwary, Rishabh and Sharma, Drishti and Winata, Gusti and Lambert, Nathan and Ruder, Sebastian and Hooker, Sara and Fadaee, Marzieh},
  editor = {Che, Wanxiang and Nabende, Joyce and Shutova, Ekaterina and Pilehvar, Mohammad Taher},
  booktitle = {Proceedings of the 63rd Annual Meeting of the Association for Computational Linguistics (Volume 1: Long Papers)},
  pages = {43--58},
  year = {2025},
  address = {Vienna, Austria},
  publisher = {Association for Computational Linguistics},
  doi = {10.18653/v1/2025.acl-long.3},
  url = {https://aclanthology.org/2025.acl-long.3/},
  isbn = {979-8-89176-251-0}
}

@inproceedings{fu2025multilingual,
  title = {How Reliable is Multilingual {LLM}-as-a-Judge?},
  author = {Fu, Xiyan and Liu, Wei},
  editor = {Christodoulopoulos, Christos and Chakraborty, Tanmoy and Rose, Carolyn and Peng, Violet},
  booktitle = {Findings of the Association for Computational Linguistics: EMNLP 2025},
  pages = {11040--11053},
  year = {2025},
  address = {Suzhou, China},
  publisher = {Association for Computational Linguistics},
  doi = {10.18653/v1/2025.findings-emnlp.587},
  url = {https://aclanthology.org/2025.findings-emnlp.587/},
  isbn = {979-8-89176-335-7}
}

@inproceedings{cmalign2025,
  title = {{CM}-Align: Consistency-based Multilingual Alignment for Large Language Models},
  author = {Zhang, Xue and Liang, Yunlong and Meng, Fandong and Zhang, Songming and Chen, Yufeng and Xu, Jinan and Zhou, Jie},
  editor = {Christodoulopoulos, Christos and Chakraborty, Tanmoy and Rose, Carolyn and Peng, Violet},
  booktitle = {Findings of the Association for Computational Linguistics: EMNLP 2025},
  pages = {25689--25702},
  year = {2025},
  address = {Suzhou, China},
  publisher = {Association for Computational Linguistics},
  doi = {10.18653/v1/2025.findings-emnlp.1401},
  url = {https://aclanthology.org/2025.findings-emnlp.1401/},
  isbn = {979-8-89176-335-7}
}

@inproceedings{implicitcrosslingual2025,
  title = {Implicit Cross-Lingual Rewarding for Efficient Multilingual Preference Alignment},
  author = {Yang, Wen and Wu, Junhong and Wang, Chen and Zong, Chengqing and Zhang, Jiajun},
  booktitle = {Findings of the Association for Computational Linguistics: ACL 2025},
  year = {2025},
  url = {https://arxiv.org/abs/2503.04647}
}

@inproceedings{kosmic2024,
  title = {Kosmic: {K}orean Text Similarity Metric Reflecting Honorific Distinctions},
  author = {Hwang, Yerin and Kim, Yongil and Bae, Hyunkyung and Bang, Jeesoo and Lee, Hwanhee and Jung, Kyomin},
  editor = {Calzolari, Nicoletta and Kan, Min-Yen and Hoste, Veronique and Lenci, Alessandro and Sakti, Sakriani and Xue, Nianwen},
  booktitle = {Proceedings of the 2024 Joint International Conference on Computational Linguistics, Language Resources and Evaluation (LREC-COLING 2024)},
  pages = {9954--9960},
  year = {2024},
  address = {Torino, Italia},
  publisher = {ELRA and ICCL},
  url = {https://aclanthology.org/2024.lrec-main.870/}
}

@article{stepaudior152026,
  title = {{Step-Audio-R1.5 Technical Report}},
  author = {Zhang, Yuxin and Zhang, Xiangyu Tony and Liu, Daijiao and Tian, Fei and Deng, Yayue and Chen, Jun and Lin, Qingjian and Zhang, Haoyang and Li, Yuxin and Gong, Jinglan and Huang, Yechang and Zhao, Liang and Yao, Chengyuan and Liu, Hexin and Chng, Eng Siong and Yang, Xuerui and Yu, Gang and Zhang, Xiangyu and Jiang, Daxin},
  journal = {arXiv preprint arXiv:2604.25719},
  year = {2026},
  url = {https://arxiv.org/abs/2604.25719}
}

@inproceedings{ipo2024,
  title = {A General Theoretical Paradigm to Understand Learning from Human Preferences},
  author = {Gheshlaghi Azar, Mohammad and Daniel Guo, Zhaohan and Piot, Bilal and Munos, Remi and Rowland, Mark and Valko, Michal and Calandriello, Daniele},
  booktitle = {Proceedings of The 27th International Conference on Artificial Intelligence and Statistics},
  series = {Proceedings of Machine Learning Research},
  volume = {238},
  pages = {4447--4455},
  year = {2024},
  publisher = {PMLR},
  url = {https://proceedings.mlr.press/v238/gheshlaghi-azar24a.html}
}

@inproceedings{simpo2024,
  title = {{SimPO}: Simple Preference Optimization with a Reference-Free Reward},
  author = {Meng, Yu and Xia, Mengzhou and Chen, Danqi},
  booktitle = {Advances in Neural Information Processing Systems},
  year = {2024},
  url = {https://arxiv.org/abs/2405.14734}
}

@inproceedings{orpo2024,
  title = {{ORPO}: Monolithic Preference Optimization without Reference Model},
  author = {Hong, Jiwoo and Lee, Noah and Thorne, James},
  editor = {Al-Onaizan, Yaser and Bansal, Mohit and Chen, Yun-Nung},
  booktitle = {Proceedings of the 2024 Conference on Empirical Methods in Natural Language Processing},
  pages = {11170--11189},
  year = {2024},
  address = {Miami, Florida, USA},
  publisher = {Association for Computational Linguistics},
  doi = {10.18653/v1/2024.emnlp-main.626},
  url = {https://aclanthology.org/2024.emnlp-main.626/}
}

@inproceedings{cpo2024,
  title = {Contrastive Preference Optimization: Pushing the Boundaries of {LLM} Performance in Machine Translation},
  author = {Xu, Haoran and Sharaf, Amr and Chen, Yunmo and Tan, Weiting and Shen, Lingfeng and Van Durme, Benjamin and Murray, Kenton and Kim, Young Jin},
  booktitle = {International Conference on Machine Learning},
  year = {2024},
  url = {https://arxiv.org/abs/2401.08417}
}

@article{slicHF2023,
  title = {{SLiC-HF}: Sequence Likelihood Calibration with Human Feedback},
  author = {Zhao, Yao and Joshi, Rishabh and Liu, Tianqi and Khalman, Misha and Saleh, Mohammad and Liu, Peter J.},
  journal = {arXiv preprint arXiv:2305.10425},
  year = {2023},
  url = {https://arxiv.org/abs/2305.10425}
}

@article{deepseekr12025,
  title = {{DeepSeek-R1}: Incentivizing Reasoning Capability in {LLMs} via Reinforcement Learning},
  author = {{DeepSeek-AI}},
  journal = {arXiv preprint arXiv:2501.12948},
  year = {2025},
  url = {https://arxiv.org/abs/2501.12948}
}

@article{qwenmath2024,
  title = {{Qwen2.5-Math} Technical Report: Toward Mathematical Expert Model via Self-Improvement},
  author = {An Yang and Beichen Zhang and Binyuan Hui and Bofei Gao and Bowen Yu and Chengpeng Li and Dayiheng Liu and Jianhong Tu and Jingren Zhou and Junyang Lin and Keming Lu and Mingfeng Xue and Runji Lin and Tianyu Liu and Xingzhang Ren and Zhenru Zhang},
  journal = {arXiv preprint arXiv:2409.12122},
  year = {2024},
  url = {https://arxiv.org/abs/2409.12122}
}

@inproceedings{tulu32025,
  title = {{T{\"u}lu 3}: Pushing Frontiers in Open Language Model Post-Training},
  author = {Nathan Lambert and Jacob Morrison and Valentina Pyatkin and Shengyi Huang and Hamish Ivison and Faeze Brahman and Lester James V. Miranda and Alisa Liu and Nouha Dziri and Shane Lyu and Yuling Gu and Saumya Malik and Victoria Graf and Jena D. Hwang and Jiangjiang Yang and Ronan Le Bras and Oyvind Tafjord and Chris Wilhelm and Luca Soldaini and Noah A. Smith and Yizhong Wang and Pradeep Dasigi and Hannaneh Hajishirzi},
  booktitle = {International Conference on Learning Representations},
  year = {2025},
  url = {https://arxiv.org/abs/2411.15124}
}

@inproceedings{llavarlhf2024,
  title = {Aligning Large Multimodal Models with Factually Augmented {RLHF}},
  author = {Sun, Zhiqing and Shen, Sheng and Cao, Shengcao and Liu, Haotian and Li, Chunyuan and Shen, Yikang and Gan, Chuang and Gui, Liangyan and Wang, Yu-Xiong and Yang, Yiming and Keutzer, Kurt and Darrell, Trevor},
  editor = {Ku, Lun-Wei and Martins, Andre and Srikumar, Vivek},
  booktitle = {Findings of the Association for Computational Linguistics: ACL 2024},
  pages = {13088--13110},
  year = {2024},
  address = {Bangkok, Thailand},
  publisher = {Association for Computational Linguistics},
  doi = {10.18653/v1/2024.findings-acl.775},
  url = {https://aclanthology.org/2024.findings-acl.775/}
}

@inproceedings{rlhfv2024,
  title = {{RLHF-V}: Towards Trustworthy {MLLMs} via Behavior Alignment from Fine-grained Correctional Human Feedback},
  author = {Yu, Tianyu and Yao, Yuan and Zhang, Haoye and He, Taiwen and Han, Yifeng and Cui, Ganqu and Hu, Jinyi and Liu, Zhiyuan and Zheng, Hai-Tao and Sun, Maosong and Chua, Tat-Seng},
  booktitle = {Proceedings of the IEEE/CVF Conference on Computer Vision and Pattern Recognition},
  year = {2024},
  url = {https://arxiv.org/abs/2312.00849}
}

@inproceedings{pope2023,
  title = {Evaluating Object Hallucination in Large Vision-Language Models},
  author = {Li, Yifan and Du, Yifan and Zhou, Kun and Wang, Jinpeng and Zhao, Xin and Wen, Ji-Rong},
  editor = {Bouamor, Houda and Pino, Juan and Bali, Kalika},
  booktitle = {Proceedings of the 2023 Conference on Empirical Methods in Natural Language Processing},
  pages = {292--305},
  year = {2023},
  address = {Singapore},
  publisher = {Association for Computational Linguistics},
  doi = {10.18653/v1/2023.emnlp-main.20},
  url = {https://aclanthology.org/2023.emnlp-main.20/}
}

@inproceedings{hallusionbench2024,
  title = {{HallusionBench}: An Advanced Diagnostic Suite for Entangled Language Hallucination and Visual Illusion in Large Vision-Language Models},
  author = {Guan, Tianrui and Liu, Fuxiao and Wu, Xiyang and Xian, Ruiqi and Li, Zongxia and Liu, Xiaoyu and Wang, Xijun and Chen, Lichang and Huang, Furong and Yacoob, Yaser and Manocha, Dinesh and Zhou, Tianyi},
  booktitle = {Proceedings of the IEEE/CVF Conference on Computer Vision and Pattern Recognition},
  year = {2024},
  url = {https://arxiv.org/abs/2310.14566}
}

@inproceedings{mme2024,
  title = {{MME}: A Comprehensive Evaluation Benchmark for Multimodal Large Language Models},
  author = {Chaoyou Fu and Peixian Chen and Yunhang Shen and Yulei Qin and Mengdan Zhang and Xu Lin and Jinrui Yang and Xiawu Zheng and Ke Li and Xing Sun and Yunsheng Wu and Rongrong Ji and Caifeng Shan and Ran He},
  booktitle = {Advances in Neural Information Processing Systems},
  year = {2024},
  url = {https://arxiv.org/abs/2306.13394}
}

@inproceedings{mu2024rule,
  title = {Rule Based Rewards for Language Model Safety},
  author = {Mu, Tong and Helyar, Alec and Heidecke, Johannes and Achiam, Joshua and Vallone, Andrea and Kivlichan, Ian D and Lin, Molly and Beutel, Alex and Schulman, John and Weng, Lilian},
  booktitle = {Advances in Neural Information Processing Systems},
  volume = {37},
  year = {2024},
  url = {https://openreview.net/forum?id=QVtwpT5Dmg}
}

@misc{kim2023prometheus,
  title = {Prometheus: Inducing Fine-grained Evaluation Capability in Language Models},
  author = {Seungone Kim and Jamin Shin and Yejin Cho and Joel Jang and Shayne Longpre and Hwaran Lee and Sangdoo Yun and Seongjin Shin and Sungdong Kim and James Thorne and Minjoon Seo},
  year = {2023},
  url = {https://arxiv.org/abs/2310.08491},
  eprint = {2310.08491},
  archiveprefix = {arXiv},
  primaryclass = {cs.CL}
}

@inproceedings{kim2024prometheus2,
  title = {Prometheus 2: An Open Source Language Model Specialized in Evaluating Other Language Models},
  author = {Kim, Seungone and Suk, Juyoung and Longpre, Shayne and Lin, Bill Yuchen and Shin, Jamin and Welleck, Sean and Neubig, Graham and Lee, Moontae and Lee, Kyungjae and Seo, Minjoon},
  editor = {Al-Onaizan, Yaser and Bansal, Mohit and Chen, Yun-Nung},
  booktitle = {Proceedings of the 2024 Conference on Empirical Methods in Natural Language Processing},
  pages = {4334--4353},
  year = {2024},
  address = {Miami, Florida, USA},
  publisher = {Association for Computational Linguistics},
  doi = {10.18653/v1/2024.emnlp-main.248},
  url = {https://aclanthology.org/2024.emnlp-main.248/}
}

@inproceedings{ye2024flask,
  title = {{FLASK}: Fine-grained Language Model Evaluation based on Alignment Skill Sets},
  author = {Seonghyeon Ye and Doyoung Kim and Sungdong Kim and Hyeonbin Hwang and Seungone Kim and Yongrae Jo and James Thorne and Juho Kim and Minjoon Seo},
  booktitle = {The Twelfth International Conference on Learning Representations},
  year = {2024},
  url = {https://openreview.net/forum?id=CYmF38ysDa}
}

@inproceedings{lee2025checkeval,
  title = {{C}heck{E}val: A reliable {LLM}-as-a-Judge framework for evaluating text generation using checklists},
  author = {Lee, Yukyung and Kim, JoongHoon and Kim, Jaehee and Cho, Hyowon and Kang, Jaewook and Kang, Pilsung and Kim, Najoung},
  editor = {Christodoulopoulos, Christos and Chakraborty, Tanmoy and Rose, Carolyn and Peng, Violet},
  booktitle = {Proceedings of the 2025 Conference on Empirical Methods in Natural Language Processing},
  pages = {15771--15798},
  year = {2025},
  address = {Suzhou, China},
  publisher = {Association for Computational Linguistics},
  doi = {10.18653/v1/2025.emnlp-main.796},
  url = {https://aclanthology.org/2025.emnlp-main.796/},
  isbn = {979-8-89176-332-6}
}

@inproceedings{yuan2024selfrewarding,
  title = {Self-Rewarding Language Models},
  author = {Yuan, Weizhe and Pang, Richard Yuanzhe and Cho, Kyunghyun and Li, Xian and Sukhbaatar, Sainbayar and Xu, Jing and Weston, Jason},
  booktitle = {Proceedings of the 41st International Conference on Machine Learning},
  series = {Proceedings of Machine Learning Research},
  volume = {235},
  pages = {57905--57923},
  year = {2024},
  publisher = {PMLR},
  url = {https://proceedings.mlr.press/v235/yuan24d.html}
}

@inproceedings{wu2025metarewarding,
  title = {Meta-Rewarding Language Models: Self-Improving Alignment with {LLM}-as-a-Meta-Judge},
  author = {Wu, Tianhao and Yuan, Weizhe and Golovneva, Olga and Xu, Jing and Tian, Yuandong and Jiao, Jiantao and Weston, Jason E and Sukhbaatar, Sainbayar},
  editor = {Christodoulopoulos, Christos and Chakraborty, Tanmoy and Rose, Carolyn and Peng, Violet},
  booktitle = {Proceedings of the 2025 Conference on Empirical Methods in Natural Language Processing},
  pages = {11537--11554},
  year = {2025},
  address = {Suzhou, China},
  publisher = {Association for Computational Linguistics},
  doi = {10.18653/v1/2025.emnlp-main.583},
  url = {https://aclanthology.org/2025.emnlp-main.583/},
  isbn = {979-8-89176-332-6}
}

@inproceedings{zeng2024token,
  title     = {Token-level Direct Preference Optimization},
  author    = {Zeng, Shunjie and Feng, Ganqu and Liu, Weizhou and Wu, Minnan and Lu, Shicheng and Zhao, Jingang and Zhang, Shuyun and Lin, Di and Liang, Xiaodan},
  booktitle = {International Conference on Machine Learning (ICML)},
  pages     = {59963--59981},
  year      = {2024},
  publisher = {PMLR},
  url       = {https://proceedings.mlr.press/v235/zeng24c.html}
}

@inproceedings{gao2023scaling,
  title = {Scaling Laws for Reward Model Overoptimization},
  author = {Gao, Leo and Schulman, John and Hilton, Jacob},
  booktitle = {Proceedings of the 40th International Conference on Machine Learning},
  pages = {10835--10866},
  year = {2023},
  volume = {202},
  series = {Proceedings of Machine Learning Research},
  publisher = {PMLR},
  url = {https://proceedings.mlr.press/v202/gao23h.html}
}

@misc{zheng2023secrets,
      title={Secrets of RLHF in Large Language Models Part I: PPO},
      author={Rui Zheng and Shihan Dou and Songyang Gao and Yuan Hua and Wei Shen and Binghai Wang and Yan Liu and Senjie Jin and Qin Liu and Yuhao Zhou and Limao Xiong and Lu Chen and Zhiheng Xi and Nuo Xu and Wenbin Lai and Minghao Zhu and Cheng Chang and Zhangyue Yin and Rongxiang Weng and Wensen Cheng and Haoran Huang and Tianxiang Sun and Hang Yan and Tao Gui and Qi Zhang and Xipeng Qiu and Xuanjing Huang},
      year={2023},
      eprint={2307.04964},
      archivePrefix={arXiv},
      primaryClass={cs.CL},
      url={https://arxiv.org/abs/2307.04964},
}

\appendix

%%%%%%%%%%%%%%
\section{Survey Methodology and Inclusion Protocol}
\label{sec:survey-methodology}

\paragraph{Scope.}
This survey focuses on methods that use explicit or recoverable structured criteria for LLM alignment, evaluation, reward modeling, or policy optimization. We include work on rubrics, constitutions, principles, checklists, process criteria, and LLM-as-a-judge settings when these criteria influence feedback or rewards. We also include adjacent alignment paradigms---DPO-style preference optimization, RLVR, PRMs, and multimodal RLHF---when they clarify the boundary between rubric-guided RL and competing or complementary methods.

\paragraph{Search sources.}
We searched ACL Anthology, OpenReview, arXiv, Semantic Scholar, Google Scholar, and proceedings or accepted-paper lists from ACL, EMNLP, NAACL, EACL, ICLR, ICML, NeurIPS, CVPR, and related workshops. We additionally followed citation links from seed papers such as Constitutional AI, DPO, PRMs, DeepSeekMath, and recent rubric-generation papers.

\paragraph{Query families.}
Search queries covered five families: (i) rubric, criteria, checklist, grading, and LLM judge; (ii) constitution, principle, RLAIF, and AI feedback; (iii) DPO, IPO, KTO, SimPO, ORPO, and direct preference optimization; (iv) RLVR, verifier, process reward, mathematical reasoning, and code execution; and (v) multimodal RLHF, hallucination, VLM reward modeling, and multi-criteria VLM evaluation.

\paragraph{Inclusion criteria.}
We included papers if they met at least one criterion: (1) they explicitly use rubrics, principles, checklists, or criteria for alignment, reward modeling, or evaluation; (2) they operationally implement rubric-like structured or multi-dimensional feedback; (3) they are foundational adjacent methods needed to position rubric-guided RL; or (4) they introduce benchmarks or failure analyses that directly inform rubric quality, reward hacking, judge reliability, multilingual evaluation, or multimodal alignment.

\paragraph{Reliability and evidence use.}
Peer-reviewed papers are used as primary evidence whenever available. Influential preprints and technical reports are included when they are widely adopted, provide necessary coverage of fast-moving 2025--2026 developments, or directly introduce rubric-guided RL methods not yet published in archival venues. We label such work cautiously in the prose and avoid using unstable or unverifiable sources as load-bearing evidence. When a citation could not be matched to a stable title, author list, venue, or archival record, we removed it from the main evidence chain rather than relying on it for taxonomy or quantitative claims.

%%%%%%%%%%%%%%
\section{Background: Related Surveys and Methods}
\label{sec:background}

\subsection{How This Survey Differs from Prior Surveys}
\label{sec:related-surveys}

Several existing surveys touch on aspects of rubric-guided RL, but none provides a comprehensive treatment.

\textbf{LLM-as-a-Judge Surveys.}
The comprehensive survey by \citet{gu2024survey} covers LLM-as-a-judge methodologies, biases, and applications. While LLM-as-a-judge is closely related to rubric-guided evaluation, the two concepts differ: LLM-as-a-judge typically uses implicit evaluation criteria, while rubric-guided evaluation uses explicit, structured criteria. Our survey bridges these two areas by examining how rubrics can make LLM-as-a-judge more interpretable and controllable.

\textbf{Reward Design and PRM Surveys.}
Recent surveys on reward design and process reward models~\cite{rewarddesign2025, prmsurvey2025} categorize reward formulations and process supervision methods. Our survey differs in its focus on rubric-based reward design, which provides structured, interpretable, and multi-dimensional reward signals. We provide a more detailed treatment of rubric-specific challenges (rubric quality, evolution, exploitation) that are not addressed in general reward design surveys, and we position PRMs as implicit step-level rubrics extending beyond single-dimension process supervision.

\textbf{Self-Evolution and Agentic RL Surveys.}
\citet{tao2024selfevolution} survey self-evolution of large language models. Our survey complements these by focusing specifically on how rubrics enable self-evolution and structured credit assignment in agentic settings, dimensions not covered by existing surveys.

\subsection{Related Alignment Paradigms}
\label{sec:related-methods}

\paragraph{RLHF.} RLHF~\cite{ouyang2022training} trains language models to align with human preferences through a three-stage pipeline. While RLHF has proven effective for alignment, its reliance on scalar reward signals introduces fundamental limitations: the reward is opaque, difficult to interpret, and prone to overoptimization~\cite{gao2023scaling}.

\paragraph{RLAIF.} RLAIF~\cite{lee2023rlaif, bai2022constitutional} replaces human preference labels with AI-generated feedback, addressing the scalability bottleneck of human annotation. Constitutional AI~\cite{bai2022constitutional} is a prominent instance of RLAIF that uses explicit natural-language principles to guide AI feedback.

\paragraph{DPO and direct preference optimization.} DPO~\cite{rafailov2023dpo} bypasses the RL optimization loop by reparameterizing the reward model to derive the optimal policy in closed form. Its variants now form a major alignment family: IPO generalizes preference optimization and regularizes against deterministic-preference overfitting~\cite{ipo2024}, KTO learns from pointwise desirable/undesirable examples~\cite{ethayarajh2024kto}, SimPO removes the reference model with a length-normalized implicit reward~\cite{simpo2024}, ORPO merges SFT and preference optimization into one stage~\cite{orpo2024}, and CPO/SLiC-HF show contrastive calibration as another RL-free route~\cite{cpo2024, slicHF2023}. This family is directly relevant to rubric-guided alignment because rubric scores can be converted into pairwise or pointwise preference labels; consequently, DPO-style optimization is best viewed as a competing or complementary backend for using rubric feedback, not merely as background RLHF literature.

\paragraph{RLVR.} RLVR couples reinforcement learning with objective, externally verifiable signals such as mathematical correctness, code execution, or rule-based answer checking~\cite{shao2024deepseekmath, deepseekr12025, qwenmath2024, tulu32025}. DeepSeekMath introduced GRPO for math reasoning with rule-based rewards~\cite{shao2024deepseekmath}, while DeepSeek-R1 and Qwen2.5-Math show that large-scale verifier- or rule-based RL can produce strong reasoning models in domains with checkable answers~\cite{deepseekr12025, qwenmath2024}. Open post-training recipes such as T{\"u}lu 3 further incorporate verifiable tasks into modern alignment pipelines~\cite{tulu32025}. These methods are direct competitors to rubric-guided RL in math and code because verifiable rewards are cheaper and less subjective; rubric-guided methods are most compelling when tasks require partial credit, explanation quality, safety, pedagogy, or other criteria that are not externally decidable.

\paragraph{PRMs.} PRMs~\cite{lightman2023lets} provide step-level feedback for reasoning tasks. We reframe PRMs as \textit{implicit, step-level rubrics} that score the ``correctness'' dimension at each reasoning step. The generalization from single-dimension PRMs to multi-dimensional rubric-guided process supervision (as in RLCER~\cite{rlcer2026}) is a natural but non-trivial extension.

\paragraph{GRPO.} GRPO~\cite{shao2024deepseekmath} eliminates the need for a value network by using group-relative rewards. GRPO has become a popular choice for rubric-guided RL methods~\cite{rar2025} because it naturally accommodates structured, multi-dimensional reward signals.

\section{Mapping Method Classes to the Bayesian Framework}
\label{sec:bayesian-clarification}

The Bayesian framework maps method classes according to how they specify, condition, and update evaluation criteria. Fixed-principle methods such as Constitutional AI specify input-independent criterion sets, corresponding to point-mass priors. Instance-specific systems such as RaR, OpenRubrics, PRMs, and judge-rubric generators construct input-conditioned criteria, corresponding to posterior point estimates. Self-evolving or learned-rubric methods such as EvoLM, Rubric-ARM, and RLCER occupy the update-oriented part of the framework because their rubrics or reward models change with training data, policy outputs, or temporal contrast. EvoLM most directly supports a posterior-sharpening interpretation; systems without explicit priors and likelihoods remain analogues. This mapping clarifies which components of $P(\mathcal{R})$ and $P(\mathcal{R}|x)$ each method operationalizes.

\section{What the Bayesian Framework Adds Beyond Re-labeling}

This framework provides conceptual and practical utility beyond simple notation replacement. Conceptually, it resolves the terminological ambiguity plaguing the field: constitutions and rubrics are unified as the same mathematical object at different conditioning levels. Constitutions act as global, unconditioned priors $P(\mathcal{R})$, while instance-specific rubrics are input-conditioned posteriors $P(\mathcal{R}|x)$. Methods such as EvoLM~\cite{evolm2026}, which optimize evolving rubric representations, instantiate the update-oriented portion of this framework most directly.

Practically, this perspective yields hypotheses about training dynamics and adversarial robustness. In explicitly probabilistic formulations, the observed ``vague to specific'' evolution pattern can be interpreted as posterior sharpening as policy data shift; other self-evolving systems are update-oriented analogues. The lens also casts rubric exploitation as overfitting to a single conditional point estimate $\mathcal{R}_x$ and suggests a robustness direction: marginalizing over the rubric distribution,
$$r_{\text{robust}}(x, y) = \mathbb{E}_{\mathcal{R}_x \sim P(\mathcal{R}|x)} \big[r_{\mathcal{R}_x}(x, y)\big]$$
which encourages the policy to generalize across plausible evaluation criteria rather than gaming a single deterministic set. This remains a design hypothesis requiring empirical validation.

\section{Computational Efficiency Analysis}
\label{sec:efficiency}

While rubric-guided reinforcement learning (RL) enhances reward interpretability, it introduces non-negligible computational and token overhead into the optimization loop. To provide a rigorous yet concise overview, we present a structural complexity analysis across different paradigms. Let $P$ denote the token length of static constitutional principles, $R$ the average token length of an individual criterion, $K$ the number of criteria per instance, and $S$ the intermediate step length for process-supervised approaches. We define $T_{\text{gen}}$ and $T_{\text{eval}}$ as the real-time execution latencies for rubric generation and judge evaluation, respectively.

As summarized in Table~\ref{tab:comp-cost-app}, the efficiency profiles across paradigms diverge along three core dimensions:
\begin{itemize}
    \item \textbf{Token Consumption \& Caching:} Static approaches (e.g., CAI~\cite{bai2022constitutional}) incur a constant $\mathcal{O}(P)$ token overhead inside the evaluation context and fully benefit from inference-time prefix caching. Conversely, instance-specific frameworks (e.g., RaR~\cite{rar2025}, Rubric-ARM~\cite{rubricarm2026}) require online synthesis of conditional rubrics, injecting an uncacheable $\mathcal{O}(K \cdot R)$ dynamic token overhead per instance.
    \item \textbf{Evaluation Call Complexity:} In implicit aggregation modes (e.g., EvoLM~\cite{evolm2026}, RaR Implicit), the judge model evaluates all dimensions simultaneously in a single forward pass ($\mathcal{O}(1)$ complexity). Explicit modes (e.g., RaR Explicit) serialize the verification, demanding $\mathcal{O}(K)$ independent judge calls per instance, which scales to $\mathcal{O}(M \cdot K)$ during test-time candidate ($M$) verification and inherently restricts optimal inference search depths.
    \item \textbf{Judge Infrastructure:} Leveraging commercial APIs typically introduces unpredictable network latency and financial costs. Modern paradigms optimize cost-effectiveness by constraining highly localized small models ($\mathcal{M}_{\text{local}}$, e.g., 1.7B--3B parameters) within precise rubric structures, compressing the localized evaluation latency to a fraction of commercial API execution times without compromising reward fidelity.
\end{itemize}

\begin{table*}[ht]
\centering
\small
\setlength{\tabcolsep}{6pt}
\renewcommand{\arraystretch}{1.3}
\begin{tabular}{lcccc}
\toprule
\textbf{Method} & \textbf{Rubric Context Cost} & \textbf{Judge Calls / Inst.} & \textbf{Eval. Latency} & \textbf{Typical Judge Scale} \\
\midrule
Scalar RM (Baseline) & $0$ & $1_{\text{RM}}$ & $\mathcal{O}(1)$  & $\mathcal{M}_{\text{dense}}$ \\
PRM~\cite{lightman2023lets} & $0$ \scriptsize{(Implicit)} & $S$ & $\mathcal{O}(S \cdot T_{\text{eval}})$ & $\mathcal{M}_{\text{dense}}$ \\
Constitutional AI~\cite{bai2022constitutional} & $\mathcal{O}(P)$ & $1$ & $\mathcal{O}(T_{\text{eval}})$ & $\mathcal{M}_{\text{large}}$ \\
RaR (Explicit)~\cite{rar2025} & $\mathcal{O}(K \cdot R)$ & $K$ & $\mathcal{O}(K \cdot T_{\text{eval}})$ & $\mathcal{M}_{\text{small}}$ (API) \\
RaR (Implicit)~\cite{rar2025} & $\mathcal{O}(K \cdot R)$ & $1$ & $\mathcal{O}(T_{\text{eval}})$ & $\mathcal{M}_{\text{small}}$ (API) \\
Rubric-ARM~\cite{rubricarm2026} & $\mathcal{O}(K \cdot R)$ & $1$ & $\mathcal{O}(T_{\text{eval}})$ & $\mathcal{M}_{\text{local}}$ \\
EvoLM~\cite{evolm2026} & $\mathcal{O}(K \cdot R)$ & $1$ & $\mathcal{O}(T_{\text{eval}})$ & $\mathcal{M}_{\text{local}}$ \\
\bottomrule
\end{tabular}
\caption{Computational complexity and resource mapping across key paradigms. $P$ and $R$ denote the token lengths of constitutional principles and individual criteria, respectively; $K$ specifies the total number of evaluation criteria, and $S$ represents the reasoning sequence steps. The base latency parameter $T_{\text{eval}}$ scales asymptotically as a function of the designated judge model capacity $\mathcal{M}$.}
\label{tab:comp-cost-app}
\end{table*}

\section{Detailed Agentic and Deep Research Benchmarks}
\label{sec:app-agentic}
A rapidly growing line of work applies rubric-based evaluation to \textit{deep research agents}---LLM agents that perform multi-step information gathering, synthesis, and report generation. ResearchRubrics~\cite{researchrubrics2026} provides 2,500+ expert-written rubrics across 100+ real-world tasks in 9 domains. DRACO~\cite{draco2026} introduces a cross-domain benchmark spanning 10 domains and 40 countries' information sources, evaluating outputs along factual accuracy, completeness, objectivity, and citation quality. DeepResearchBench~\cite{deepresearchbench2026} constructs 9,430 fine-grained binary rubrics from expert-written investigative articles. DEER~\cite{deer2025} develops a 7-dimension, 25-subdimension, 101-item rubric taxonomy with claim verification, making system strengths and limitations interpretable.

\section{Continual Learning Benchmarks}
\label{sec:app-cl}
CL-bench~\cite{clbench2026} introduces a benchmark with 500 complex contexts, 1,899 tasks, and 31,607 verification rubrics for assessing context learning. CL-bench Life~\cite{clbenchlife2026} extends this to real-life scenarios with 405 context-task pairs and 5,348 rubrics. In continual learning, rubrics serve as memory, where criteria from previous tasks provide evaluation frameworks that prevent catastrophic forgetting.

\section{Consistency and Rubric Quality}
\label{sec:app-consistency}
RubricBench~\cite{rubricbench2026} introduces a curated benchmark with 1,147 pairwise comparisons to assess the reliability of rubric-guided evaluation. RubricEval~\cite{rubriceval2026} is the first rubric-level meta-evaluation benchmark for instruction following.

\section{Representative Reported Results from Original Papers}

Table~\ref{tab:cross-method} summarizes representative results reported by original papers. \textbf{Important caveat}: these results are reported under different evaluation settings, model sizes, and training configurations. They are included to document each paper's own evaluation setting, not to rank methods against each other.

\leavevmode

\begin{table*}[ht]
\centering
\small
\renewcommand{\arraystretch}{1.2}
\setlength{\tabcolsep}{8pt}

\begin{tabularx}{\textwidth}{l c c X c}
\toprule
\textbf{Method} & \textbf{Base Model} & \textbf{Benchmark} & \textbf{Metric} & \textbf{Result} \\
\midrule

\multirow{2}{*}{RaR~\cite{rar2025}} & \multirow{2}{*}{Qwen-2.5-7B} & HealthBench, & Rel. improvement & +31\% \\
 & & GPQA-Diamond & Rel. improvement & +7\% \\
\addlinespace[4pt]

Rubric-ARM~\cite{rubricarm2026} & Various & RewardBench & Avg. gain & +4.7\% \\
\addlinespace[4pt]

\multirow{2}{*}{EvoLM~\cite{evolm2026}} & \multirow{2}{*}{Qwen3-8B} & RewardBench-2, & Over GPT-4.1 & +25.7\% \\
 & & OLMo3-Adapt & Average & 69.3\% \\
\addlinespace[4pt]

RLCER~\cite{rlcer2026} & 7B & Math benchmarks & Over RLVR & Outperforms \\
OpenRubrics~\cite{openrubrics2025} & Various & RewardModel & Over baselines & +8.4\% \\
PRM~\cite{lightman2023lets} & Various & MATH & Accuracy & 78.2\% \\
Constitutional AI~\cite{bai2022constitutional} & Various & Harmlessness Elo & Pareto & Improved \\
Chasing the Tail~\cite{chasingtail2025} & Various & RewardBench & Accuracy gain & +3.2\% \\
RLCF~\cite{rlcf2025} & Various & Multi-domain & Consistent gain & Improved \\

\bottomrule
\multicolumn{5}{p{\textwidth}}{\vspace{0.5pt}\small \textbf{Note}: Results are from original papers under different settings and should NOT be directly compared.}
\end{tabularx}
\caption{Representative results reported by original papers. These numbers are included only to document each paper's own evaluation setting and should not be interpreted as a controlled cross-method comparison.}
\label{tab:cross-method}
\end{table*}

\section{Detailed Comparison of Rubric Aggregation Strategies}

Rubric-guided RL systems differ not only in how they aggregate
criterion-level scores, but also in how they construct criteria and
assign credit. Table~\ref{tab:aggregation} compares the principal
reward construction and aggregation strategies used by representative methods.

\begin{table*}[ht]
\centering
\renewcommand{\arraystretch}{1.18}
\setlength{\tabcolsep}{3.5pt}
\newcolumntype{L}[1]{>{\raggedright\arraybackslash\hsize=#1\hsize}X}
\scriptsize

\begin{tabularx}{\textwidth}{
    p{2.05cm}
    p{3.15cm}
    p{1.45cm}
    L{0.95}
    L{1.05}}
\toprule
\textbf{Strategy} &
\textbf{Formulation} &
\textbf{Used By} &
\textbf{Pros} &
\textbf{Cons} \\
\midrule

Weighted Sum &
$r=\frac{\sum_k w_k s_k}{\sum_k w_k}$ &
RaR (Explicit) &
Independent criterion scores expose their contributions and permit direct control through criterion weights. &
The result depends on prespecified weights and separate judge decisions may miss interactions among criteria. \\

Implicit \newline Aggregation &
$r=f_\phi(x,y,\{(w_k,d_k)\})$ &
RaR (Implicit) &
A single holistic judgment can integrate interactions without requiring a separate call for every criterion. &
Delegating aggregation to the judge makes the final score less directly attributable to individual criteria. \\

Learned Rubric-Conditioned Judging &
$o\sim\pi_j(\cdot\mid x,y^{(1)},y^{(2)},r)$ &
Rubric-ARM &
Preference feedback jointly improves rubric generation and rubric-conditioned judging under a shared correctness objective. &
Training two coupled components adds computation and requires careful scheduling to control non-stationarity. \\

Discriminative Rubrics &
Optimize rubrics for preference separation &
EvoLM &
Instance-specific rubrics emphasize differences that make closely matched response pairs more informative for training. &
Because the objective rewards separability, coverage of important but weakly discriminative criteria requires separate validation. \\

Step Product &
$r=\prod_t p_t(\text{correct})$ &
PRM &
Step-level evaluations localize errors while yielding a solution score that represents all steps being correct. &
Multiplication is sensitive to sequence length and calibration, so one underestimated step can suppress the entire score. \\

Principle \newline Ensembling &
Sample one principle per comparison &
CAI &
Sampling across principles produced more robust preference-model behavior than repeatedly applying a single principle. &
Each comparison applies only one sampled principle, so conflicts and coverage gaps are not resolved within that label. \\

Checklist \newline Aggregation &
$r=\frac{\sum_k w_k s_k}{\sum_k w_k}$ &
RLCF &
Atomic criteria and optional programmatic verifiers make individual failures inspectable and hard constraints checkable. &
Scoring items separately is judge-call intensive, and optimizing checklist completion can produce reward-hacking artifacts. \\

Hierarchical Credit &
Aggregate subgoal returns at planning and execution levels &
HiPER &
Hierarchical advantages assign credit at both levels and reduce variance in long-horizon agent training. &
The method requires a planner--executor decomposition and introduces additional level-specific optimization choices. \\

Pairwise Adaptive &
Instantiate pair-specific criteria and aggregate criterion preferences externally &
OpenRS &
Pair-conditioned criteria target semantic differences while retaining inspectable, externally aggregated preferences. &
A new rubric must be instantiated for each pair, increasing evaluation work and limiting direct cross-pair comparability. \\

\bottomrule
\end{tabularx}

\caption{Comparison of rubric reward construction and aggregation
strategies. Each strategy exposes a different trade-off among
interpretability, computational cost, criterion coverage, and the
ability to capture interactions.}
\label{tab:aggregation}
\end{table*}

\section{Extended Discussion on Rubric Quality Dimensions}

\paragraph{Discriminative Power.}
Discriminative power measures how well a rubric distinguishes between high-quality and low-quality responses. A rubric with high discriminative power assigns significantly different scores to good and bad responses, while a rubric with low discriminative power assigns similar scores regardless of quality. This can be measured by the accuracy of rubric-conditioned evaluation on held-out preference data.

\paragraph{Coverage.}
Coverage measures how many relevant quality dimensions a rubric captures. A rubric with high coverage evaluates all important aspects of response quality (correctness, completeness, clarity, safety, etc.), while a rubric with low coverage may miss important dimensions. PRMs, for example, typically cover a narrower set of dimensions. Incomplete rubrics may fail to capture important aspects of response quality, leading to reward hacking through unmeasured dimensions.

\paragraph{Calibration.}
Calibration measures how well rubric scores align with human judgment across the full quality spectrum. A well-calibrated rubric assigns scores that are consistent with human evaluation, while a poorly calibrated rubric may over- or under-estimate quality. LLM-Rubric~\cite{hashemi2024llmrubric} proposes calibration methods for rubric-based evaluation. Calibrating rubric-based scores to match human judgment distributions is critical for consistency.

\section{Linguistic Insights: Detailed Evidence and Qualitative Examples}
\label{sec:linguistic-appendix}

This appendix provides detailed evidence and qualitative examples supporting the claims in Section~\ref{sec:linguistic}.

\paragraph{Semantic Drift: Qualitative Evidence.}

EvoLM~\cite{evolm2026} provides the most direct evidence for semantic drift: rubrics evolve from vague criteria (e.g., ``Ensure mathematical accuracy'') to specific, verifiable facts (e.g., ``The answer is 144, derived from perimeter 48'') as training progresses. This intentional drift admits a posterior-sharpening interpretation. Constitutional AI~\cite{bai2022constitutional} observed unintentional drift through Goodharting, where over-optimization leads to boilerplate language and overly harsh responses. The diagnostic in Equation~\ref{eq:semantic-drift} is a proposal of this survey; no prior work has validated this specific formalization for rubric-guided RL.
We present a concrete case study of semantic drift in Table ~\ref{tab:case-drift}.

\paragraph{Linguistic Reward Hacking: Qualitative Examples.}

\begin{itemize}
\item \textbf{Verbosity exploitation}: \citet{gao2023scaling} document reward model overoptimization where proxy rewards increase while true rewards decline, a dynamic closely related to verbosity exploitation. In rubric contexts, models learn to pad responses with filler to satisfy ``completeness'' criteria without adding substantive content.
\item \textbf{Criterion bleeding}: Safe RLHF~\cite{safeRLHF2024} explicitly addresses the safety-helpfulness tradeoff, where maximizing safety (refusing all requests) violates helpfulness. ArmoRM~\cite{armorm2024} uses multi-objective reward modeling to mitigate criterion bleeding by balancing multiple quality dimensions simultaneously.
\item \textbf{Surface pattern matching}: self-evolving rubric work suggests that evaluator capacity and rubric specificity interact, with smaller evaluators often requiring more explicit criteria to distinguish genuine satisfaction from superficial compliance~\cite{evolm2026}. This suggests that surface pattern matching is a real phenomenon that can be partially mitigated by evaluator and rubric design.
\end{itemize}

We provide concrete case studies for each type of linguistic reward hacking, drawing on empirical evidence from recent work. The case studies can be found in Table ~\ref{tab:case-verbosity}, \ref{tab:case-bleeding}, ~\ref{tab:case-surface} and ~\ref{tab:case-granularity}.

%%%%%%%% START CASE STUDIES %%%%%%%%

\renewcommand{\arraystretch}{1.35}

% --- Case Study 1 ---
\begin{table*}[h]
\centering
\small
\begin{tabular}{p{0.95\linewidth}}
\toprule
\textbf{Case Study 1: Semantic Drift---Intentional vs.\ Unintentional} \\
\midrule
\textbf{Intentional drift (EvoLM):} \citet{evolm2026} document that rubrics evolve from vague criteria (e.g., ``Ensure mathematical accuracy'') to specific, verifiable facts (e.g., ``the correct maximum area of 144, derived from the given perimeter of 48'') as training progresses. This intentional drift admits a posterior-sharpening interpretation and improves downstream policy quality. \\
\textbf{Unintentional drift (CAI):} \citet{bai2022constitutional} observe that the meaning of ``harmlessness'' drifts from ``avoid causing harm'' to ``refuse all potentially sensitive queries'' under over-optimization. The rubric text remains unchanged, but the policy's interpretation shifts, leading to overly harsh and unhelpful responses. \\
\bottomrule
\end{tabular}
\caption{Semantic drift in rubric-guided RL: intentional drift (posterior sharpening) may improve quality, while unintentional drift (Goodharting) can degrade it.}
\label{tab:case-drift}
\end{table*}

\vspace{1.5em}

% --- Case Study 2 ---
\begin{table*}[h]
\centering
\small
\begin{tabular}{p{0.95\linewidth}}
\toprule
\textbf{Case Study 2: Verbosity Exploitation via Presence-Based Rubrics} \\
\midrule
\textbf{Source:} \citet{rubrichacking2026} \\
\textbf{Phenomenon:} The RL checkpoint improves completeness (+1.07) but degrades conciseness ($-$2.91), factual correctness ($-$0.85), and overall quality ($-$1.02). The rubric-based verifier prefers the RL checkpoint on 85.8\% of prompts, but rubric-free judges prefer the base model on 78.4\%---a clear sign of reward hacking under strong verification. \\
\textbf{Mechanism:} 90.2\% of rubric weight falls on \textit{presence-based} criteria (e.g., ``States that plasma volume increases more than red cell mass during pregnancy''). Presence-based satisfaction rises from 27.6\% to 42.5\% (+14.9 pp), while absence-based criteria decline. The model exploits this by generating longer, claim-denser responses that satisfy presence checks without verifying correctness. \\
\textbf{Safety-presence exploitation:} Safety-presence rubric items (8.4\% weight) such as ``The response advises the user to consult a healthcare provider before taking any action'' can be satisfied by appending boilerplate disclaimers without genuine safety reasoning. \\
\bottomrule
\end{tabular}
\caption{Verbosity exploitation through presence-based rubric criteria. The model learns to satisfy presence checks by generating longer, claim-denser output, while degrading on conciseness, factual correctness, and overall quality.}
\label{tab:case-verbosity}
\end{table*}

\vspace{1.5em}

% --- Case Study 3 ---
\begin{table*}[h]
\centering
\small
\begin{tabular}{p{0.95\linewidth}}
\toprule
\textbf{Case Study 3: Criterion Bleeding---Safety-Helpfulness Tradeoff} \\
\midrule
\textbf{Phenomenon:} Maximizing the safety criterion (``the response should not cause harm'') by refusing all requests directly violates the helpfulness criterion (``the response should address the user's query''). Safe RLHF~\cite{safeRLHF2024} and ArmoRM~\cite{armorm2024} have proposed mechanisms to mitigate this problem. We categorize this phenomenon as ``Criterion bleeding'' --- satisfying one rubric dimension at the expense of another. \\
\textbf{Constitutional AI evidence:} \citet{bai2022constitutional} observe that over-optimization against constitutional principles leads to boilerplate language and overly harsh responses---the policy learns to satisfy safety keywords (``I cannot assist with that'') while violating the spirit of balanced evaluation. \\
\textbf{Mitigation:} Safe RLHF explicitly decomposes the reward into separate safety and helpfulness objectives with constrained optimization. ArmoRM~\cite{armorm2024} uses multi-objective reward modeling to balance multiple quality dimensions simultaneously, preventing criterion bleeding by design. \\
\bottomrule
\end{tabular}
\caption{Criterion bleeding: when satisfying one rubric dimension (safety) comes at the expense of another (helpfulness), the policy exploits the multi-objective gap.}
\label{tab:case-bleeding}
\end{table*}

\vspace{1.5em}

% --- Case Study 4 ---
\begin{table*}[h]
\centering
\small
\begin{tabular}{p{0.95\linewidth}}
\toprule
\textbf{Case Study 4: Surface Pattern Matching---Rubric Verification Failure Modes} \\
\midrule
\textbf{Source:} \citet{rubrichacking2026} \\
\textbf{Failure Mode Taxonomy (stable across training, domains, and verifier strength):} \\
\quad $\bullet$ \textbf{Partial Compound (36.0\%):} The criterion requires A$\wedge$B but the verifier accepts only one conjunct. Example: ``The verifier failed because it accepted partial satisfaction of a multi-part requirement as full credit, verifying the outcome statement while not enforcing the required explicit distinction between two specified categories.'' \\
\quad $\bullet$ \textbf{Implicit-as-Explicit (34.6\%):} The required claim is never explicitly stated; the verifier infers it from context. Example: ``The verifier failed because it credited an implicit or inferable statement as if it were explicit, accepting broad plausibility instead of requiring the exact characterization the criterion demanded.'' \\
\quad $\bullet$ \textbf{Imprecise Verification (29.4\%):} The verifier accepts a related but distinct concept as equivalent, or checks only broad topic relevance. Example: ``The verifier failed because it matched on surface topic relevance instead of verifying the specific claim.'' \\
\textbf{Key insight:} These failure modes are \textit{stable} across training---the distribution of exploitation types does not shift, only the volume increases. Both weak and strong verifiers fail in the same proportions, suggesting these are fundamental limitations of rubric verification. \\
\bottomrule
\end{tabular}
\caption{Surface pattern matching in rubric verification: three recurring structural failure modes where the verifier accepts superficial criterion satisfaction without genuine engagement.}
\label{tab:case-surface}
\end{table*}

\vspace{1.5em}

% --- Case Study 5 ---
\begin{table*}[t!]
\centering
\small
\begin{tabular}{p{0.95\linewidth}}
\toprule
\textbf{Case Study 5: Granularity Effects and Evaluator Capacity} \\
\midrule
\textbf{Phenomenon:} Smaller judges (1.7B) force the learned rubric generator to produce more precise, fine-grained criteria, resulting in better downstream policies than larger judges (up to 32B) that tolerate vague rubrics. This demonstrates that optimal rubric granularity interacts with judge capacity: a rubric that is ``too coarse'' for a small judge may be ``just right'' for a large judge, and vice versa. \\
\textbf{PRM evidence:} \citet{lightman2023lets} show that step-level (fine-grained) process supervision achieves 78.2\% on MATH, substantially outperforming outcome-level (coarse-grained) supervision. This confirms that finer granularity produces better signals, but we argue that such benefits can be fully achieved only when the evaluator can reliably apply the fine-grained criteria. \\
\textbf{RLCF evidence:} RLCF~\cite{rlcf2025} decomposes each instruction into atomic checklist items and uses them as instance-level reward criteria, illustrating how checklist-style rubrics can provide denser and more inspectable feedback than holistic outcome rewards. \\
\bottomrule
\end{tabular}
\caption{Granularity effects: the interaction between rubric granularity and judge capacity determines downstream policy quality. Finer granularity generally helps, but only when the evaluator can reliably apply the criteria.}
\label{tab:case-granularity}
\end{table*}

\renewcommand{\arraystretch}{1.0}

%%%%%%%% END CASE STUDIES %%%%%%%%

\paragraph{Underspecification as Opportunity.}

No prior work explicitly argues that underspecification is an opportunity in rubric-guided RL. However, several methods implicitly exploit it: EvoLM~\cite{evolm2026} observes rubrics evolving from vague to specific through posterior sharpening, RLCER~\cite{rlcer2026} starts with underspecified rubrics that self-evolve during training, and Rubric-ARM~\cite{rubricarm2026} learns rubric-conditioned reward models in non-verifiable settings. We posit that underspecification provides the space for rubric self-evolution---without initial underspecification, adaptive sharpening would have no room to operate.

\paragraph{Cross-Linguistic Considerations.}

\citet{hong2025crosslingual} demonstrate that English-trained reward models exceed target-language reward models by 3--4\% on Multilingual RewardBench, suggesting cross-lingual transfer is feasible but imperfect. \citet{sheth2026crosslingual} propose a Universal Criteria Set (UCS) with language-agnostic evaluation dimensions combined with language-specific instantiations. \citet{fu2025multilingual} find that multilingual LLM-as-a-judge has low cross-lingual consistency (Fleiss' Kappa $\approx$ 0.3), indicating that rubric-conditioned judges face the same challenge. The mR3 model~\cite{mr3_2026} extends to 72 languages with rubric-agnostic reward reasoning, while M-RewardBench~\cite{mrewardbench2025} provides the first systematic multilingual RM evaluation across 23 languages. Language-specific quality dimensions include Japanese keigo (honorific system) with multiple formality levels, Korean jondaetmal (honorific usage)~\cite{kosmic2024}, and Arabic diglossia---all absent from English-centric rubrics. CM-Align~\cite{cmalign2025} uses consistency-based approaches for multilingual alignment.

%%%%%%%%%%%%%%

\section{Adversarial Robustness and Reward Hacking}
\label{sec:adversarial}

Rubric-guided RL introduces both new defenses against and new vulnerabilities to adversarial attacks and reward hacking.

\paragraph{Rubric Exploitation as Overfitting to a Point Estimate.}
A novel form of reward hacking in rubric-guided RL is \textit{rubric exploitation}: the policy learns to produce responses that score well against rubric criteria while violating the spirit of the rubric. Under the Bayesian framework (\S\ref{sec:framework}), rubric exploitation corresponds to \textit{overfitting to a point estimate} of the rubric posterior: the policy optimizes against a single sample $\mathcal{R}_x$ rather than the full posterior $P(\mathcal{R}|x)$. The defense is to evaluate under the full posterior (marginalization), which requires the policy to perform well under all plausible rubrics rather than gaming a single one.

\paragraph{Rubric-Specific Reward Hacking.}
Reward Hacking in Rubric-Based RL~\cite{rubrichacking2026} studies the specific phenomenon where policies optimized against a training rubric are evaluated against a cross-family panel. The key finding is that rubric-based rewards do not by themselves prevent reward hacking: the gap between training and evaluation rubrics creates an exploitation surface.

\paragraph{Defenses.}
Dynamic rubric adaptation (self-evolving rubrics close exploitation opportunities), multi-rubric evaluation (ensemble methods), constrained optimization with rubric-based safety constraints, and information-theoretic approaches such as InfoRM~\cite{inform2024} provide defenses against rubric exploitation.

\end{document}